\documentclass[lettersize,journal]{IEEEtran}
\usepackage{amsmath,amsfonts}
\DeclareMathOperator*{\argmax}{arg\,max}
\usepackage{algpseudocode}
\usepackage{algorithm}
\usepackage{array}
\usepackage[caption=false,font=normalsize,labelfont=sf,textfont=sf]{subfig}
\usepackage{textcomp}
\usepackage{stfloats}
\usepackage{url}
\usepackage{verbatim}
\usepackage{graphicx}
\usepackage{cite}
\usepackage{xcolor}
\usepackage{orcidlink}
\newcommand{\etal}{\textit{et al}. }
\newcommand{\ie}{\textit{i}.\textit{e}., }
\newcommand{\eg}{\textit{e}.\textit{g}. }

\newcommand\red[1]{{{#1}}}

\begin{document}

\title{AnyGrasp: Robust and Efficient Grasp Perception in Spatial and Temporal Domains}

\author{Hao-Shu Fang\orcidlink{0000−0002−0758−0293}, Chenxi Wang\orcidlink{0000-0002-9420-3495}, Hongjie Fang\orcidlink{0000-0002-6309-1160}, Minghao Gou\orcidlink{0000-0003-1425-4231},\\ Jirong Liu\orcidlink{0000-0002-8444-5677}, Hengxu Yan\orcidlink{0000−0003−4860−2652}, Wenhai Liu\orcidlink{0000-0003-0166-7774}, Yichen Xie\orcidlink{0000−0003−4443−8795}, Cewu Lu\orcidlink{0000−0002−0758−0293},~\IEEEmembership{Member,~IEEE}
\thanks{H-S. Fang, C. Wang, H. Fang, M. Gou, J. Liu, H. Yan, W. Liu, Y. Xie and C. Lu are with the School of Electronic Information and Electrical Engineering, Shanghai Jiao Tong University, Shanghai, 200240, China (email: fhaoshu@gmail.com, \{wcx1997, galaxies, gmh2015, jirong, hengxuyan, sjtu-wenhai, xieyichen, lucewu\}@sjtu.edu.cn). C.Lu is the corresponding author.}
}

\markboth{IEEE Transactions on Robotics}%
{Fang \MakeLowercase{\textit{et al.}}}

\IEEEpubid{0000--0000/00\$00.00~\copyright~2022 IEEE}
\maketitle
\begin{abstract}
    As the basis for prehensile manipulation, it is vital to enable robots to grasp as robustly as humans. Our innate grasping system is prompt, accurate, flexible, and continuous across spatial and temporal domains. Few existing methods cover all these properties for robot grasping. In this paper, we propose AnyGrasp for grasp perception to enable robots these abilities \red{using a parallel gripper}. Specifically, we develop a dense supervision strategy with real perception and analytic labels in the spatial-temporal domain. Additional awareness of objects' center-of-mass is incorporated into the learning process to help improve grasping stability. Utilization of grasp correspondence across observations enables dynamic grasp tracking. Our model can efficiently generate accurate, 7-DoF, dense, and temporally-smooth grasp poses and works robustly against large depth-sensing noise. Using AnyGrasp, we achieve a 93.3\% success rate when clearing bins with over 300 unseen objects, which is on par with human subjects under controlled conditions. Over 900 mean-picks-per-hour is reported on a single-arm system. For dynamic grasping, we demonstrate catching swimming robot fish in the water.
\end{abstract}


\section{Introduction}

\IEEEPARstart{V}{isual} guided object grasping is important in the robotic community. Other than solely picking objects in the industrial environment, we would foresee robots working closely with humans even in household environments. Thus, a grasping system that can serve most daily manipulation in unstructured environments is attractive.

To achieve that, we first introspect our own grasping ability. For humans, we perceive a partial observation without a full model of the scene. Our visual system processes such input within 100ms and we would then know how to pre-shape our hand for comforting contact during grasping~\cite{baldauf2010attentional}. Such a grasp system is accurate and robust for any object. Moreover, a time-consistent property is guaranteed so that we can recognize and grasp moving objects in dynamic scenes. 

To date, researchers have proposed different methods for visual-guided grasping. Some methods~\cite{traditional1, traditional3} assumed full knowledge of the object and contact model, which does not always hold in the real world. Some simplified the grasping perception as a planar detection problem~\cite{2014redmon, dexnet2} but would impose restrictions on afterward manipulation. Some proposed sampling-evaluation methodologies sample candidates from the scene and then evaluate their quality~\cite{gpd, dexnet2, dexnet4, mousavian20196}. However, this approach is time-consuming and cannot generate dense predictions. Furthermore, these methods mainly focus on static scene grasp detection, and the dynamic grasp detection problem \red{remains largely unexplored}.

In this paper, we present a spatially and temporally unified methodology, AnyGrasp, to bridge the grasp perception ability gap between robots and humans. \red{We focus on grasping with a parallel gripper in this paper.} Specifically, a geometry processing module estimates dense 7-DoF grasp configurations for a monocular perceived observation in one feed-forward pass. A temporal association module identifies the grasp correspondence among these tremendous grasp poses across every two observations. To reduce the burden of explicit collision detection during grasping, our model is obstacle-aware and eliminates the grasp candidates that have no space for hand placing. This helps accelerate the selection of grasp poses in afterward manipulation. To improve stability, the awareness of the center of gravity (COG) for objects is also equipped. These two properties are also witnessed in human's visually guided grasping behavior~\cite{baldauf2010attentional}. Our model can generate accurate, 7-DoF and continuous grasp poses across space and time in 100ms. 

A problem during learning is that we lack a dataset for dynamic object grasping in the real world. Previous grasp learning resorts to collecting data in simulation. However, we show in this paper that when using a low-cost commercial depth sensor, an algorithm with simple sim-to-real transferring techniques still performs worse than directly training with real-world data. Thus, we persist in training with real data. To avoid extensive human labor, we exploit the grasp pose correspondence across different observations of a static scene and propose a dense supervision strategy with real perception and analytic labels in the spatial-temporal domain that greatly improves data efficiency. 

\IEEEpubidadjcol

We have conducted relevant research~\cite{graspnet, gou2021rgb, wang2021graspness} before. However, we only considered the grasping problem in static scenes and did not fully evaluate the grasp perception system in different scenarios. Thus, in this paper, we develop the grasping ability in dynamic scenes that is complementary to our previous work. \red{Besides, we did not clearly show the advantages of training with real-world data~\cite{graspnet} over simulated data. In this paper, we explicitly demonstrate the superiority of our grasping system, trained with a relatively small amount of real-world objects (144 objects), on a variety of challenging tasks that are not presented in previous work trained with thousands of simulated objects.} Extra analyses of different principles in dataset design are also conducted. \red{We hope our discovery will encourage the community in related areas to prioritize real-world training data.}

Specifically, we first test our algorithm in a large-scale bin-picking experiment with over 300 unseen objects with diverse shapes, materials, and sizes. \red{Our results demonstrate an average success rate of over 93\% and a completion rate of over 99.8\%, which is comparable to the performance of humans using the same end-effector configuration and open-loop grasping strategy.} We further carry out experiments of our method on different depth sensing noise to show its robustness. For dynamic scenes, the robot successfully demonstrates catching moving robot fish in the fish tank, which is difficult due to the small friction in the water and the tiny size of the fish. Finally, we discuss the influence of several factors in dataset construction, such as real-world data versus simulated data, the number of grasp annotations on an object, the scene diversity, etc. These analyses may be helpful for future data collection in this area.

\red{The novelty and contribution of this paper include:
\begin{itemize}
    \item We propose the first unified system for fast, accurate, 7-DoF and temporally-continuous grasp pose detection, using a parallel gripper.
    \item We incorporate the awareness of object COG for grasp perception and propose a new generation-association methodology for dynamic 7-DoF grasp configuration prediction.
    \item We demonstrate the robustness of our method, trained on real-world data consisting of \textit{only 144 objects}, through extensive experiments in many challenging scenarios.
    \item We release a grasping library that has on-par performance with human subjects tested on over 300 unseen objects, with over 900 mean picks per hour (MPPH).
    \item We provide a detailed analysis of different training factors, such as the selection of real or simulated data, the influence of annotation density, the importance of scene variance, etc.
\end{itemize}
}
To support reproducible research, the library and example code of AnyGrasp system is provided in supplementary materials.

\section{Related work}
In this section, we introduce some background material that most relates to the visual grasping topic. The scope is restricted to the two-finger parallel gripper, which we mainly discuss in this paper. This section is divided into two parts, where Sec.~\ref{sec:grasp_detection} focuses on grasp pose detection methods and Sec.~\ref{sec:continuous_action} focuses on continuous action learning methods.

\subsection{Grasp Pose Detection}
\label{sec:grasp_detection}
The grasp pose detection problem can be defined as predicting several poses for a given scene, usually in the Cartesian space, such that the robot can successfully lift the objects when moving its end-effector to those poses. Early methods would assume full 2D or 3D knowledge of the objects~\cite{bicchi2000robotic, shimoga1996robot}, or approximate the objects into a set of primitive shapes~\cite{miller2003automatic, bowers2003manipulation}. These methods would meet limitations in real-world environments where the 3D models of objects are hard to obtain. Learning-based methods help to alleviate this dilemma through large-scale data and automatic feature extraction. Representations for grasp pose detection evolved with the development of learning-based methods, including a point-based representation~\cite{saxena2006robotic, saxena2008robotic, saxena2008learning},  a point pair representation~\cite{le2010learning}, a rectangle representation~\cite{jiang2011efficient, lenz2015deep, multi, redmon2015real, pinto2016supersizing, dexnet1, dexnet2}, a grasp quality maps representation~\cite{morrison2018closing}, etc. These methods mainly generate 4 DoF grasp poses on the camera plane. The limitation in \red{the} degree of freedom may neglect some vital grasp poses, \eg grasp poses on the edge of a plate, and result in failure.

 To generate the full 6 DoF grasp poses, Ten Pas~\etal\cite{gpd} proposed a sampling-evaluation-based method that firstly samples grasp candidates on point cloud and then evaluates their grasp qualities with a neural network. Different sampling or evaluation models are later proposed~\cite{liang2019pointnetgpd, shao2020unigrasp, mousavian20196}. A major drawback of the sampling-evaluation methods is that they need to trade off the computation time and the number of generated grasp poses. Thus, they usually took several seconds to run and only generated tens of grasp poses for a scene. Recently, Fang~\etal\cite{graspnet} proposed an end-to-end network that directly generates abundant grasp poses for an input scene point cloud. A large-scale dataset with real data and analytic labels was also built. Similar end-to-end networks were parallelly developed in~\cite{qin2020s4g, zhao2021regnet, sundermeyer2021contact}. \red{The major difference between these methods is how they represent the $SE(3)$ grasp pose.} Later, Wang~\etal\cite{wang2021graspness} proposed a graspness module that eliminates unfeasible grasp poses in the early stage of the network, which greatly improves the grasp success rate. Other than point cloud, some researchers also explored other representations of a scene, including RGB/D image~\cite{gou2021rgb, dong2022graspvdn} and neural radiance field~\cite{ichnowski2021dex}. A recent survey~\cite{newbury202newbury2022review2deep} provided a full summary of the 6 DoF grasping methods. In this paper, we follow the end-to-end methodology and mainly build our network upon~\cite{wang2021graspness}. \red{To improve grasp stability, we further incorporate awareness of the object's center-of-mass into the learning process, which has been less considered before.}

The problem formulation of grasp pose detection constrains them to focus on static scenes. In order to extend to the dynamic scene, classic methods usually require prior information about the objects~\cite{kim2014catching, menon2014motion, akinola2021dynamic}, or operate on a fixed set of grasp trajectories~\cite{marturi2019dynamic}. For dynamic grasping without prior, existing methods either choose the nearest grasp pose in the next image as the same grasp target~\cite{morrison2018closing} or generate possible future candidates and evaluate their grasp quality~\cite{yang2021reactive}. Both methods only guarantee the tracked grasp pose has a small distance with respect to the last frame in the image coordinate system, but cannot guarantee a small distance in the object coordinate system. \red{In this paper, we propose a new generation-association methodology for predicting dynamic 7-DoF grasp configurations.}

\subsection{Continuous Action Learning for Grasping}
\label{sec:continuous_action}
Different from the above methods that detect the grasp poses in the scene and move toward the target grasp pose using a motion planner, this line of research directly maps the observation to a continuous action space. \red{This line of research has a different problem formulation than our paper, so we can only provide a summary of some representative work due to space limitations.} Levine~\etal\cite{levine2018learning} collected 800K grasp attempts on real robots and learn a CNN grasp predictor that associates image and motor command. Viereck~\etal\cite{viereck2017learning} proposed to learn with simulated point clouds, such that no real-robot trials are needed. Bousmalis~\etal\cite{bousmalis2018using} investigated the sim-to-real transfer problem in grasp action learning. Song~\etal\cite{song2020grasping} collected grasping demonstrations with a low-cost hand-held gripper. Wang~\etal\cite{wang2022goal} extended the continuous action space to 6D. A problem of these methods is that they may be prone to the domain shift problem~\cite{ross2011reduction, tobin2017domain} and perform less robust than the detection-based methods.

\red{\subsection{Training Data for 6-DoF Grasping}
For the 6-DoF grasping problem, various simulated object sets have been collected, including DexNet~\cite{dexnet1}, which has over 1000 objects, EGAD~\cite{morrison2020egad}, which generates over 2000 objects using an evolutionary algorithm, and Acronym~\cite{eppner2021acronym}, which collects over 8000 objects. With each passing year, the number of objects in these datasets continues to increase. However, it remains unclear how many training objects are actually necessary, and the increasing number of objects may make the training process increasingly burdensome. In this paper, we demonstrate that a real-world dataset consisting of only 144 objects can provide a model with grasp performance comparable to that of human subjects. We hope that our findings will encourage the community to identify critical objects for the grasping problem.
}

\section{AnyGrasp design principles}
In this section, we first describe the problem definition of our spatial-temporal grasp detection. Then we introduce the principles of our system design, from both the algorithm and data perspectives.
\subsection{Problem Definition}
For parallel jaw-based grasping, we represent a grasp pose $\mathcal{G}$ as
\begin{equation}
    \mathcal{G} = [\mathbf{R}\ \mathbf{t}\ w],
\end{equation}
where $\mathbf{R} \in \mathbb{R}^{3\times3}$ denotes the gripper orientation, $\mathbf{t} \in \mathbb{R}^{3\times1}$ denotes the center of grasp, and $w \in \mathbb{R}$ denotes the minimum gripper width that is suitable for grasping the target object. \red{This representation covers all the degrees of freedom for a parallel gripper and is also referred to as the 7-DoF grasp configuration.} \red{In this paper, We use the notation $\mathcal{E}$ to represent the environment including the robot and objects, $\mathcal{P}$ to represent the partial-view point cloud from a depth camera, and $s(\mathcal{E}, \mathcal{P}, \mathcal{G})$ to denote a binary variable indicating the success or failure of grasp $\mathcal{G}$ given environment $\mathcal{E}$ and perception $\mathcal{P}$.} A grasp is successful if the object is lifted successfully. Our goal is to find a set of grasp poses $\mathbf{G}=\left\{\mathcal{G}_1,\mathcal{G}_2,\cdots, \mathcal{G}_n\right\}$ that maximizes the grasp success rate given a fixed $n$:
\begin{equation}
    \{\mathcal{G}^{*}_1,\mathcal{G}^{*}_2,\cdots, \mathcal{G}^{*}_n\} = \argmax_{|\mathbf{G}|=n} \sum_{\mathcal{G}_i \in \mathbf{G}} \text{Prob}(s=1| \mathcal{E}, \mathcal{P}, \mathcal{G}_i ).
\label{eqn:graspposes}
\end{equation}
This means that we hope our algorithm predicts abundant grasp poses to cover the whole scene so that we can have different candidates for grasp execution.

When introducing the temporal dimension, we denote $\mathcal{E}^{t}, \mathcal{P}^{t}, \mathbf{G}^{t}$ as the environment, perception and grasp poses at time $t$, $dist(\mathcal{G}_k^{t}, \mathcal{G}_k^{t-1} | \mathcal{E}^{t}, \mathcal{E}^{t-1})$ as the distance of the paired grasp poses across two moments under the coordinate system of the grasped target. Thus, we would have:
\begin{equation}
\begin{aligned}
    &\left\{\mathcal{G}^{*}_1,\mathcal{G}^{*}_2,\cdots, \mathcal{G}^{*}_n\right\}^{(t)} = \argmax_{|\mathbf{G}^{t}|=n} \sum_{\mathcal{G}_i \in \mathbf{G}^{t}} \text{Prob}(s=1| \mathcal{E}^{t}, \mathcal{P}^{t}, \mathcal{G}_i ),\\
    &\text{s.t.}~dist(\mathcal{G}_k^{t}, \mathcal{G}_k^{t-1} | \mathcal{E}^{t}, \mathcal{E}^{t-1}) \le \delta, \forall~\mathcal{G}_k^{t-1} \in \mathbf{G}^{t-1},
\end{aligned}
\label{eqn:temgraspposes}
\end{equation}
$\delta$ is the tolerance error, under which two grasp poses could be regarded as the same pose.

\subsection{Spatial-Continuous Learning}
Previous methods~\cite{gpd, dexnet4} adopt the sampling strategy that chooses candidates on point cloud and evaluates their quality. Instead, our geometry processing module directly perceives the single-view point cloud of the scene and estimates the grasp quality across the $\mathbb{R}^6$ space. We refer to it as ``spatial-continuous learning'' owing to the dense property of the predicted grasp poses, from which we can query a feasible grasp pose at any target location in most cases.

Compared to the sampling-based methods that classify the quality of a grasp solely based on the cropped local point cloud within the gripper space, our method also considers the geometric structure from neighbor regions that could provide richer information conveying whether a grasp is of good quality or not. Such global geometric features can be easily learned through a 3D convolutional network. By taking the whole scene as input, two extra advantages are witnessed. Firstly, to improve the stability of grasping, human tends to put visual attention on the center of gravity (COG) of the object at the preparing stage of grasping~\cite{baldauf2010attentional}. Since the mass distribution is unknown, humans usually assume the COG is the center of the object.  Such intuition can be modeled by our neural network that directly perceives the whole scene. In our network, we encode such information for each grasp pose by predicting a normalized vertical distance from the gripper plane to the COG of the grasped object. Secondly, researchers in cognition found that humans would visually attend to the obstacle during the preparation period of grasping~\cite{baldauf2010attentional, deubel2004attentional}. That is, the visual system would also consider the obstacle and avoid collision for hand pre-shaping. \red{Such ability can only emerge when the network takes the whole scene as input. This aspect has also been emphasized in previous methods that adopt end-to-end networks~\cite{graspnet, qin2020s4g, zhao2021regnet, sundermeyer2021contact}.} In our network, if there is an obstacle around the grasp pose and leaves no space for gripper pre-shaping, the grasp quality score is directly set to zero.  Although obstacle avoidance can also be performed using mesh and point cloud, the neural-based implicit collision detection helps eliminate a large portion of unreachable grasp poses and reduces the expensive computation time.

\subsection{Temporal-Continuous Learning}
When trying to grasp a moving object, the robot needs to continuously update the target grasp pose before catching the object. The grasp poses between \red{every} two frames should have \red{a} small SE(3) distance  in the object's coordinate system to enable a smooth, target-consistent movement of the robot gripper. We refer to the learning of such property as ``temporal-continuous learning''.

Previous sampling-based method~\cite{yang2021reactive} added some disturbance to the grasp pose from the previous frame and evaluate these candidates in the current frame's geometry. To ensure computation efficiency, the generated proposals are usually sparse. Thus they cannot cover all the possible movements of the objects.

To avoid the speed-accuracy tradeoff, we propose a new generation-association methodology to ensure dense and consistent grasp poses temporally. Given two observations at different times, our geometry processing module generates dense grasp poses across the scene. A temporal association module takes the grasp poses and their corresponding geometric features encoded by the spatial model as input and produces their many-to-many association score matrix. The association score between each two grasp poses denotes their consistency in the temporal domain, measured by their SE(3) distance in the grasping object's coordinate system. Contributed to the spatial continuous property of our geometry processing module, we could generate dense grasp poses around a selected target to ensure both temporal continuity and grasp quality.

\subsection{Training Data}
For data collection, most methods for high DoF grasp pose detection directly learn from simulation. The main reason is that obtaining dense grasp pose annotation for data collected in the real world is difficult. However, although turning to simulation lowers the training cost, an expensive and high-precision depth camera is needed~\cite{dexnet4} to achieve good performance during the inference phase. This is to bridge the sim-to-real gap since simulation produces perfect partial view depth. In contrast, we choose to directly learn from real-world perception, which requires more effort for data collection but enables the algorithm to adapt to real-world noise, especially on low-cost cameras. In our experiments, we show how our grasp detection system can tolerate different sensing noises, and improve the success rate by a large margin over its counterpart trained in simulation.

\section{Methods and Materials}

\subsection{Data Collection and Annotation}
\label{sec:datacollect}
For the training data of our algorithm, we mainly adopt the training set of GraspNet-1Billion~\cite{graspnet} and follow the same methodology to collect 168 extra scenes composed of 104 new objects.
In brief, we obtain objects' 3D mesh models with a commercial 3D scanner and calculate dense grasp poses using analytic antipodal scores~\cite{dexnet2,gpd}. For each scene, we randomly choose several objects ($\sim$10) and randomly placed them on the table. Images are taken at 256 different viewpoints for each scene and we manually annotate the object 6D poses in the scene \red{to ensure the accuracy}. Then the grasp poses on the scene can be obtained by projecting the poses on each object with the annotated 6D pose to the scene.  Moreover, collision detection is performed by \red{simplifying the gripper model into three cubes} and checking whether each grasp pose has an intersection with the object models in the scene. \red{In our pipeline, all annotations except for the 6D poses are automatically labeled by the program.}. For more details, we refer readers to~\cite{graspnet}. In total, 268 scenes that consist of 144 objects are used to train our network. 

In this paper, we annotate three extra labels for the training data, which help improve the stability during grasping and enable the grasp pose tracking ability.

\begin{figure}[!t]
\centering
\includegraphics[width=0.5\textwidth]{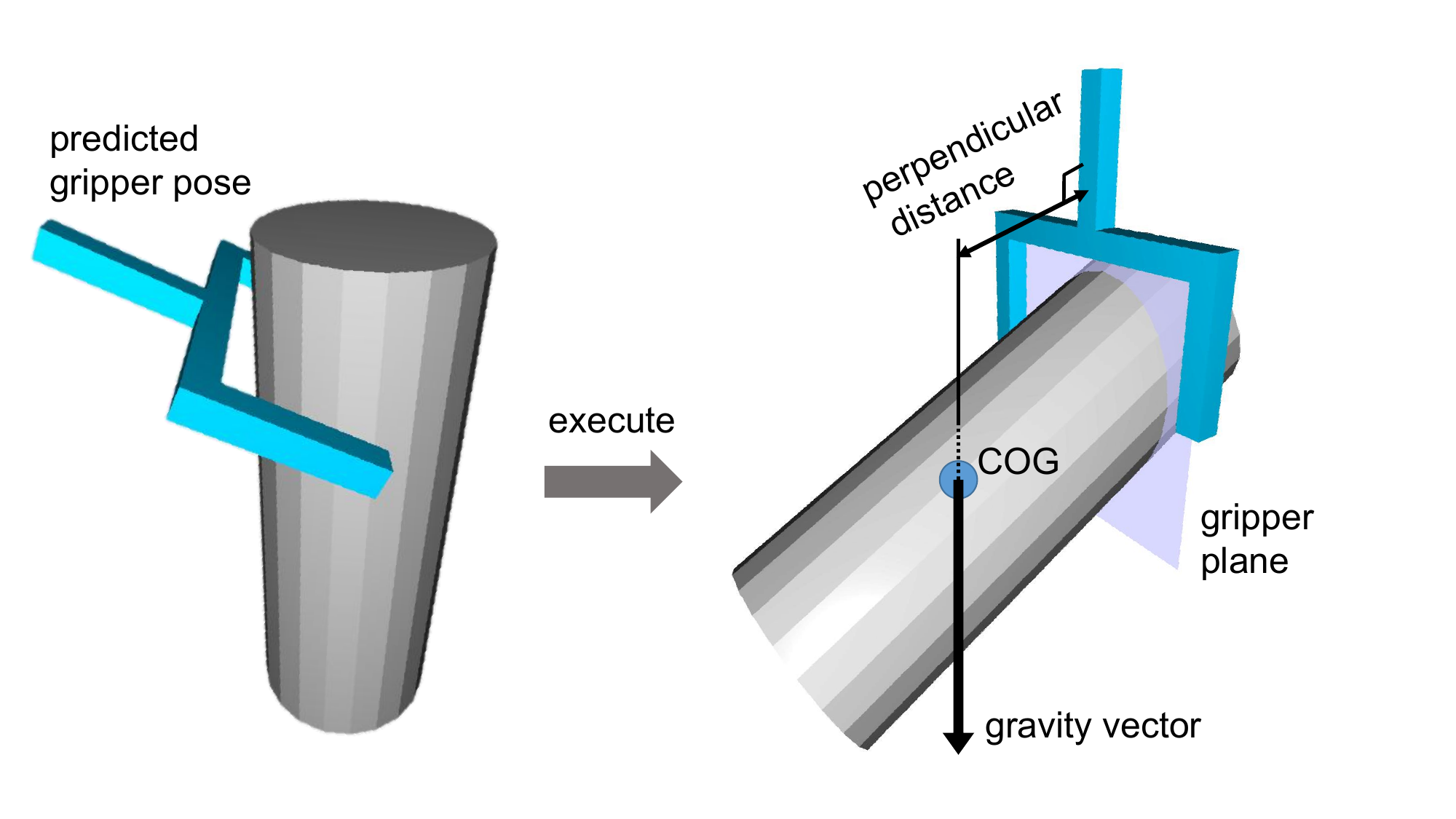}
\caption{An illustration of the perpendicular distance from the gripper plane to the COG of the object. \red{Note that we assume the gripper will move to the vertical pose, parallel to the gravity force vector when transporting the object. We observe that this policy provides a larger workspace for our robot.}}
\label{fig_cog}
\end{figure}
Firstly, the original grasp pose annotation consists of four approach depths, \ie 1, 2, 3, and 4 centimeters. In this paper, we add an extra depth of 0.5 centimeters to grasp small objects.

Secondly, as illustrated in~\cite{baldauf2010attentional}, human attention will bias to the center of gravity of the object during grasping. Inspired by this, we define a stable score for the grasp pose. We assume that the gripper will \red{move to a vertical pose that is parallel to the force of gravity when transporting the object after it is grasped}. Thus, we define the stable score as the normalized perpendicular distance between its gripper plane and the COG of the object. The normalization process is conducted by obtaining all the perpendicular distances for an object and then dividing them by the maximum distance among them. In this case, the lower this score is, the more disturbance the grasp pose can tolerate since the gravity moment is more balanced.  To obtain the annotation, we first compute the COG of the object. Since the center cannot be precisely computed only from visual perception, the object is assumed to be a solid rigid body with uniform density, which is similar to human intuition. Then, the COG is transformed into the gripper coordinate system according to the grasp pose. The perpendicular distance is the distance from the COG to the gripper plane. Fig.~\ref{fig_cog} gives an example of calculating the perpendicular distance for a grasp pose.  Finally, all the distances are collected and normalized to $[0,1]$ to get stable scores.

Thirdly, to ensure temporal consistency of the grasp poses, our algorithm associates the grasp poses that represent the same pose \textit{w.r.t} the target object between two frames. To achieve that, a training dataset that contains the association label is needed. We generate such association labels for our collected training data. Although the original dataset does not contain dynamic scenes with moving objects, it contains images captured from 256 different viewpoints for each scene. The images captured from adjacent viewpoints present subtle differences which share similar patterns when objects are moving. Thus, for each image in the training data, we first find its neighbor images with adjacent viewpoints. For each pair of adjacent images, we annotate the association of grasp poses by defining a distance metric. For two grasp poses on the same object, we transform them to the same coordinate frame (\ie the object frame) and compute two distances:
\begin{equation}
\begin{split}
    &\Delta \mathbf{R} = \arccos\frac{\mathrm{trace}(\mathbf{R}_1^\top\mathbf{R}_2)-1}{2},\\
    &\Delta \mathbf{t} = ||\mathbf{t}_1 - \mathbf{t}_2||,
\end{split}
\end{equation}
where $\mathcal{G}_1=[\mathbf{R}_1\ \mathbf{t}_1\ w_1]$ and $\mathcal{G}_2=[\mathbf{R}_2\ \mathbf{t}_2\ w_2]$ are the transformed grasp poses, $\mathrm{trace}(\cdot)$ is the trace of a matrix, $\Delta \mathbf{R}$ and $\Delta \mathbf{t}$ are the rotation distance and the translation distance respectively. The distance between two grasp poses is defined as
\begin{equation}\label{eqn:grasp-distance}
    d(\mathcal{G}_1,\mathcal{G}_2) = \frac{\Delta\mathbf{t}}{w_{\max}} + \gamma\frac{\Delta\mathbf{R}}{\pi},
\end{equation}
where $w_{\max}$ is the maximum gripper width, $\gamma$ is a distance balancing weight. In practice, we set $w_{\mathrm{max}}=0.01\mathrm{m}$ and $\gamma=0.1$. The distance between grasps on different objects is set to infinity. 

\begin{figure*}[!t]
\centering
\includegraphics[width=0.95\textwidth]{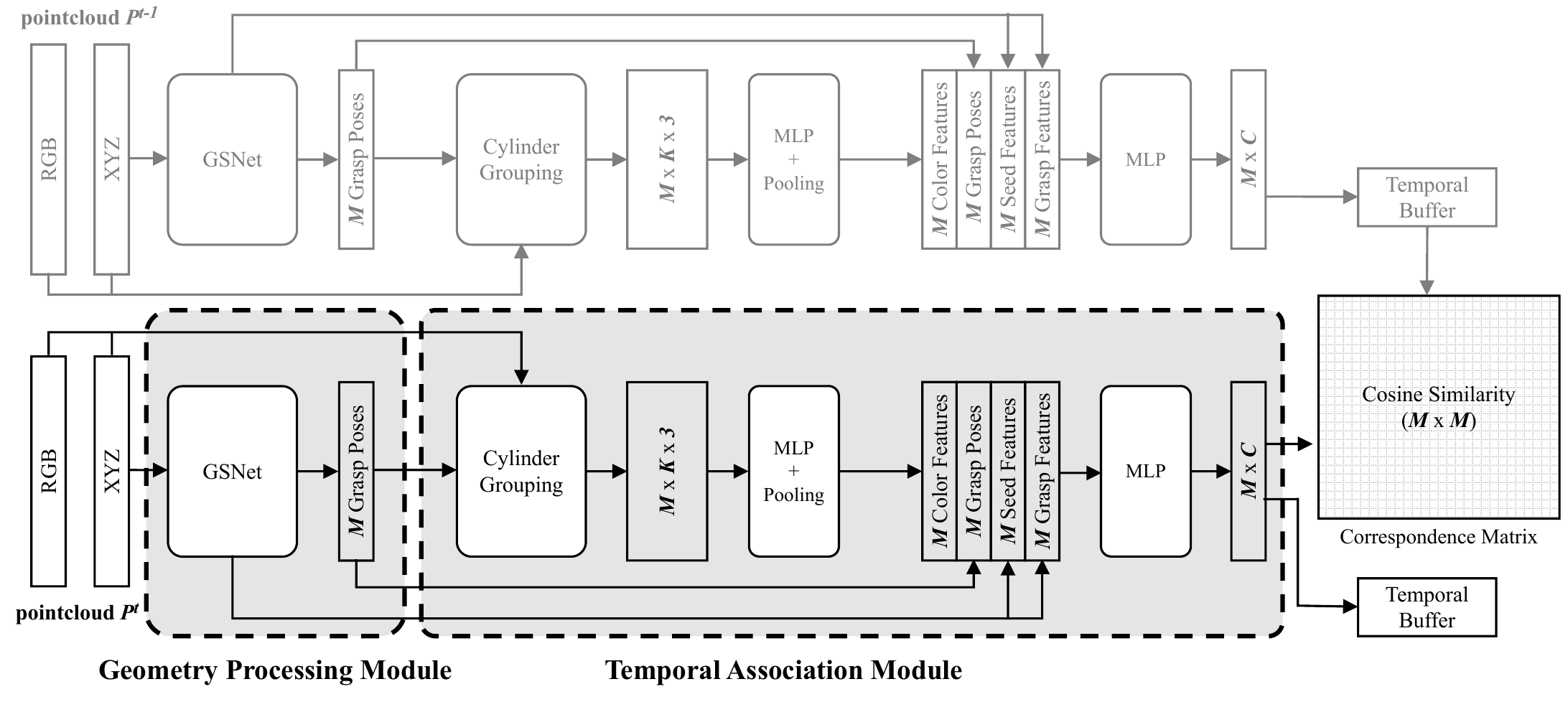}
\caption{Illustration of the model architecture for AnyGrasp. For a partial point cloud, the geometry processing module predicts dense grasp poses. The temporal association module generates \red{a feature vector with size $\mathcal{C}$} for each of the $\mathcal{M}$ predicted grasp poses. The \red{feature vector is} learned with the objective that for the grasp pose pair with a smaller distance in the object coordinate frame, the higher cosine similarity they have. Thus we can construct the correspondence matrix for the dense grasp pose pairs across two frames. Details of the model structure are given in the text.}
\label{fig_model}
\end{figure*}

\begin{figure}[!t]
\centering
\includegraphics[width=0.45\textwidth]{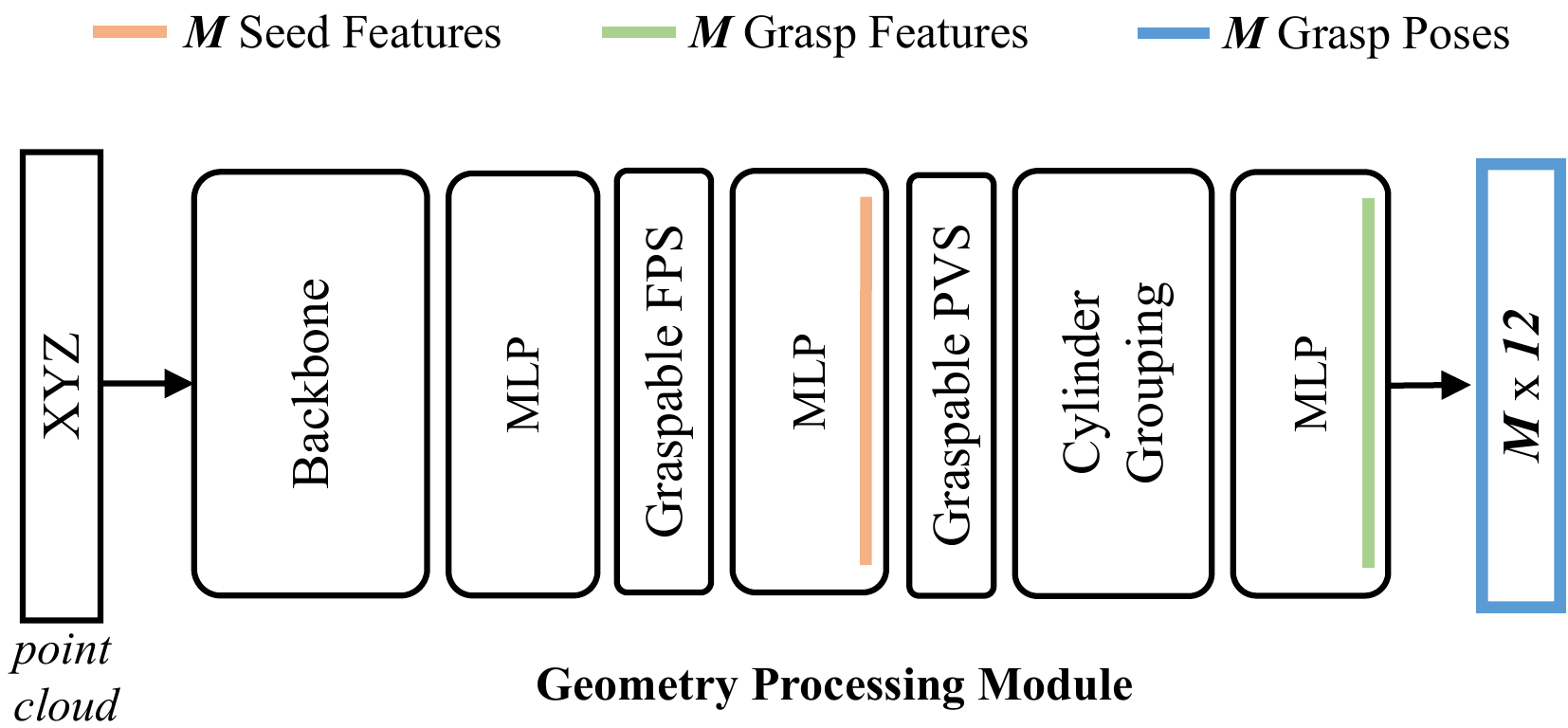}
\caption{Detailed illustration of the geometry processing module. \red{The meaning of each block like the ``Backbone'', ``MLP'', ``Graspable FPS'', etc. are specified in Sec.\ref{subsec:gsnet}.} The position of the seed features, grasp features, and grasp poses are denoted in different colors. }
\label{fig_smallmodel}
\end{figure}

\subsection{Grasp Perception Model Details}
\label{subsec:gsnet}
Next, we illustrate the details of our algorithm, which consists of the geometry processing module and the temporal association module. Fig.~\ref{fig_model} illustrates the model structure.
\subsubsection{Geometry processing module}
\red{Our geometry processing module is based on GSNet~\cite{wang2021graspness}, with a minor modification to incorporate the stable score into the network.  To explain our approach, we first provide an overview of GSNet, which is also shown in Fig.\ref{fig_smallmodel}. Instead of directly predicting $\mathbf{R},\mathbf{t}$ and $w$, GSNet decomposes these parameters into grasp point, view, in-plane rotation, approach depth, and width.} Given a point cloud \red{$\mathcal{P}$}, a 3D convolutional backbone extracts geometric features for each point. \red{A multi-layer perceptron (MLP) block generates an objectness mask and a heatmap indicating each point's graspable probability based on the extracted features.} \red{Graspable farthest point sampling (Graspable FPS) is performed to sample $\mathcal{M}$ seed points from the scene according to the objectness mask and the graspable probability map ($\mathcal{M}$ is set to 1024).} For each sampled point, \red{another MLP block} generates 300 view scores to determine \red{the most suitable view for grasping through graspable probabilistic view selection (Graspable PVS). The cylinder grouping module} groups the local geometric features along that view within a cylinder space. For each group, \red{an MLP block} predicts the grasp scores for 48 grasp poses composed of 12 in-plane rotations and 4 approach depths along that view, as well as 48 grasp widths for these grasp poses. \red{The in-plane rotation is discretized, but a heuristic search\cite{8560458} can also be used.} We refer readers to~\cite{wang2021graspness} for more details of GSNet. 

The original GSNet predicts $12 \times 4 \times 2$ values for each sampled point at the last layer, representing $12 \times 4$ grasp scores and $12 \times 4$ grasp widths for the grasp poses along the best view. We modify the last layer and the network now predicts $12 \times 5 \times 2 + 12$ values. The change from 4 to 5 is because we annotate an extra approach depth for the grasp poses, as stated in Sec.~\ref{sec:datacollect}. The extra $12$ predictions denote the stable scores for grasp poses with different angles since the stable score is shared across different grasp depths.

During inference, we multiply the original grasp score by $(1-stable\_score)$ as the new score for a grasp pose. \red{Then we reparameterize the predicted grasp point, view, in-plane rotation, approach depth, and width into a 7-DoF grasp pose $\mathcal{G} = [\mathbf{R}\ \mathbf{t}\ w]$. We then rank all predicted grasp poses according to their scores, and select the top-$n$ grasp poses as the set of grasp poses for use in Eqn.~\ref{eqn:graspposes}.}

\subsubsection{Temporal association module}
On top of the geometry processing module, we develop a temporal association module to enable grasp pose tracking. The key of the temporal association module is to generate \red{a feature vector} for each grasp pose so that we can compute the correspondence score between each two grasp poses across time.

We first introduce how we construct \red{the feature vector} for each grasp pose. Given an input point cloud, the geometry processing module outputs $\mathcal{M}$ grasp poses by choosing the pose with the highest score on each of the $\mathcal{M}$ seed points. Concurrently, we can obtain the seed features and the grasp features of the seed points. These features are extracted before the last layer of two MLP blocks in the GSNet. Fig.~\ref{fig_smallmodel} illustrates the feature location. These features mainly represent the geometric cues of the grasp poses. Considering that the texture or color information is also a strong cue for tracking, we also extract the color features of each seed point. Using the predicted grasp pose, we group the RGB information along the grasp direction using cylinder grouping, where the input seed features are replaced by seed colors. RGB information of $\mathcal{K}$ points inside the cylinder space \red{($\mathcal{K}$ is set to 16)} are grouped for each of the $\mathcal{M}$ seed points, resulting in a size of $\mathcal{M\times K \times} 3$. The local texture information is forwarded through an MLP block with pooling layers to obtain the color features. Then the color features, seed features, grasp features, and the grasp poses (parameters of shape $\mathcal{M}\times 12$, the length 12 is composed of rotation matrix (9) and translation (3)) are concatenated and fed into an MLP block to get the \red{feature vector} with size $\mathcal{C}$ for each predicted grasp \red{($\mathcal{C}$ is set to 256)}.

After obtaining the \red{feature vector} of each grasp pose, we can calculate the correspondence score between each two grasp poses. Let $\mathbf{f}_1$ be the feature vector for grasp $\mathcal{G}_1$ in the first point cloud, and $\mathbf{f}_2$ for grasp $\mathcal{G}_2$ in the second point cloud, we calculate the correspondence score using cosine similarity:
\begin{equation}
    s_{\text{corres}}(\mathcal{G}_1, \mathcal{G}_2) = \frac{\mathbf{f}_1\cdot\mathbf{f}_2}{||\mathbf{f}_1||\cdot||\mathbf{f}_2||}.
\end{equation}
The scores for all $(\mathbf{f}_1,\mathbf{f}_2)$ pairs are computed and form the correspondence matrix. 

During training, two point clouds from adjacent viewpoints are respectively forwarded through the geometry processing module and the temporal association module to obtain their \red{feature vectors} for $\mathcal{M}$ grasp poses, both having a size of $\mathcal{M}\times \mathcal{C}$. Then we calculate the correspondence matrix by the cosine similarity. The predicted correspondence matrix with size $\mathcal{M}\times \mathcal{M}$ is compared with the ground-truth matrix and the loss is back-propagated to the temporal association module. The loss function is introduced in the next subsection. 

During inference, only the point cloud at the current time is required to pass the network, and the \red{feature vectors} will be stored in a temporal buffer. The correspondence matrix is computed with the current features and those stored in the temporal buffer from the last frame. If we want to track \red{$n$} certain grasp poses selected in the previous step, we can compare its feature vector of size \red{$n\times\mathcal{C}$} with the $\mathcal{M}$ feature vectors in the current frame and generate an association vector of size \red{$n\times\mathcal{M}$}. Then we pick the feasible grasp poses with the \red{top-$n$} correspondence scores as the next prediction. \red{These grasp poses are the set of grasp poses we want in Eqn.~\ref{eqn:temgraspposes}.}

\subsubsection{Loss function}
The loss function for the geometry processing module follows~\cite{wang2021graspness}. \red{It contains softmax loss for the objectness classification in the first MLP block, and smooth-$l_1$ loss for the graspable heatmaps and grasp pose parameter regression in the three MLP blocks.} We refer readers to~\cite{wang2021graspness} for more details.

For the temporal association module, two associated grasp poses should have a high corresponding score, which indicates they have similar features. So we adopt supervised contrastive learning~\cite{khosla2020supervised} in model training, which pulls together the features from the same class.

For two grasp poses $\mathcal{G}_1$ and $\mathcal{G}_2$, they are treated as the same class if and only if $d(\mathcal{G}_1,\mathcal{G}_2)\leq\sigma$. The predicted grasp sets for two point clouds are denoted by $\mathbf{G}^1 = \{\mathcal{G}_i^1 | i=1,2,\cdots,M\}$ and $\mathbf{G}^2 = \{\mathcal{G}_j^2 | j=1,2,\cdots,M\}$ respectively. For a grasp pose $\mathcal{G}_i^1$ in the first point cloud, we collect all the grasp poses in the second point cloud belonging to the same class of $\mathcal{G}_i^1$, which are denoted by $\mathbf{P}(i)=\{\mathcal{G}_k^2\in \mathbf{G}_2 | d(\mathcal{G}_i^1,\mathcal{G}_k^2)\leq\sigma\}$. The loss function is defined as
\begin{equation}
    L = \sum_{\mathcal{G}_i^1\in \mathbf{G}_1} \frac{-1}{|\mathbf{P}(i)|} \sum_{\mathcal{G}_k^2\in P(i)}\log \frac{\exp(s_\mathrm{corres}(\mathcal{G}_i^1,\mathcal{G}_k^2)/\tau)}{\sum_{\mathcal{G}_j^2\in \mathbf{G}_2}\exp(s_\mathrm{corres}(\mathcal{G}_i^1,\mathcal{G}_j^2)/\tau)},
\end{equation}
where $|\mathbf{P}(i)|$ stands for the cardinality of $\mathbf{P}(i)$ and $\tau$ stands for the temperature parameter. \red{In our experiment, we set $\sigma=0.1$ and $\tau=0.1$.}

\subsection{Training Details}
Input point clouds are down-sampled with a voxel size of 0.005m. We set $\mathcal{M}=1024$ for each scene and \red{$\mathcal{C}=256$ for each feature vector}. \red{$\mathcal{K}$ is set to 16 in the cylinder grouping of temporal module}. The two modules are trained on the extended GraspNet-1Billion dataset using one Nvidia GTX 2080 Ti GPU with Adam optimizer~\cite{kingma2014adam} and an initial learning rate of 0.001. We adopt the ``poly'' policy with $power=0.9$ for learning rate decay, which is used in DeepLab~\cite{chen2017deeplab}.

We first train the geometry processing module from scratch with a batch size of 4. For data augmentation, we randomly flip the scene horizontally, \red{and} randomly rotate the points by Uniform$[-30^\circ,30^\circ]$ around the Z-axis (in the camera coordinate frame). We also randomly translate the points by Uniform$[-0.2,0.2]$m in X/Y-axis and Uniform$[-0.1,0.2]$m in Z-axis.

After the geometry processing module converged, we freeze its weights and train the temporal association module jointly with it. Each mini-batch contains four pairs of point clouds, where the two point clouds in one pair are captured by neighboring viewpoints from the same scene. 
Besides random flipping, random rotation, and random translation, we also randomly remove some objects in the scene for data augmentation \red{with a probability of 0.2.} 

\subsection{Detection Post-processing}
For the detected grasp poses, we conduct two post-processing \red{steps} to improve the stability. The first step is to perform collision detection. Although our network also implicitly learns whether a grasp pose would collide with the scene, such \red{an} obstacle-aware property is not a hard constraint and may be prone to noise. Thus, we perform extra collision detection for the top-100 grasp poses among the predicted results. The collision detection is based on the partial-view point cloud by examining whether there are any points within the gripper-occupied grid\red{, where the gripper is simplified into three cubes}. This step provides a safety guarantee in most cases.

The second step is to perform a gripper-centering process. It comes from an observation that if the two fingertips of the gripper contact the object surface sequentially, the earlier contact may push the object away and cause a failed grasp. The main reason is that the GraspNet-1Billion dataset does not restrict the gripper fingertips to have the same distance to the object. Thus, we perform the gripper centering process by calculating the distance from both fingertips to their contact points and translating the gripper along the connection direction of the fingertips to ensure the same moving distance from both fingertips to their contact points. \red{We define the "contact points" as the outermost points inside the gripper space. We transform the partial-view point cloud to the gripper frame based on the predicted grasp poses. It is possible that the actual contact points are not visible in the partial view. However, we found that this step ensures the gripper is center-located in most real-world cases.}

The above two processes are implemented on GPU with matrix computation and take 80 ms for 100 grasp poses. \red{During the execution phase, our method outputs the top-100 ranking grasp poses, and a self-implemented grasp planner is used to select a target grasp pose. Further details are provided in Section~\ref{sec:expproc}.}

\section{Experimental Setup}
\begin{figure*}[!t]
\centering
\includegraphics[width=0.9\textwidth]{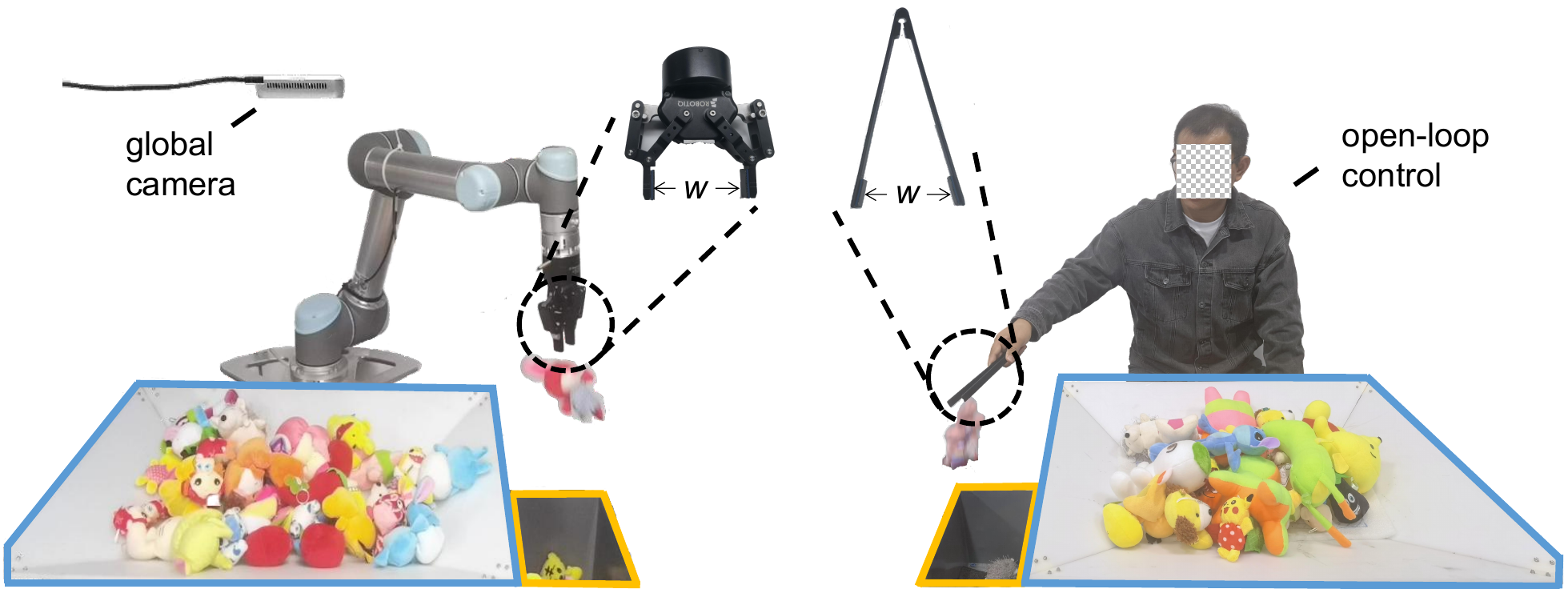}
\caption{Illustration of the hardware and volunteer setting in the static scene. Objects are moved from the object plate (surrounded by a blue line) to the target bin (surrounded by a yellow line). The robot is equipped with a global camera. We ensure the same fingertip material and the same opening width of the robot gripper and the human-held jaw. The human subjects are asked to perform an open-loop control, where adjusting the grasp pose with tactile feedback after contacting the object is not allowed.}
\label{fig_static_setting}
\end{figure*}

To verify the performance of our grasp perception system, we embed it with real robot platforms and conduct grasping experiments.
\subsection{Hardware and Human Subject}
For the static scene grasping, we use a UR5 robot arm with an overhead camera. Intel RealSense D415 and D435 are adopted to evaluate the algorithm performance across depth sensors. We use a Robotiq-85 gripper. In~\cite{dexnet4}, the authors used a customized silicone soft fingertip~\cite{guo2017design} that is designed for robust grasping. Since it is not available to us, we attach a soft table tennis rubber\footnote{\url{https://www.amazon.com/Rubbers-DHS-Table-Tennis/s?rh=n\%3A3419421\%2Cp_89\%3ADHS}} to the fingertips, which is publicly available to any research group. 

We also invite human subjects to conduct the bin picking and compare it with the AnyGrasp perception system. Human subjects are required to use a two-finger jaw for a fair comparison. The two-finger jaw has the same opening width as the robot gripper. Note that since we focus on the visual perception for grasping, we ask the volunteers to adopt an open-loop strategy during grasping. It means that they shall not adjust the grasp pose with tactile feedback after contacting the object. To enable fair comparison with human grasping, we attach the same soft rubber to the human hold jaw. Fig.~\ref{fig_static_setting} shows the robot and human subject settings.

\begin{figure}[!t]
\centering
\includegraphics[width=0.45\textwidth]{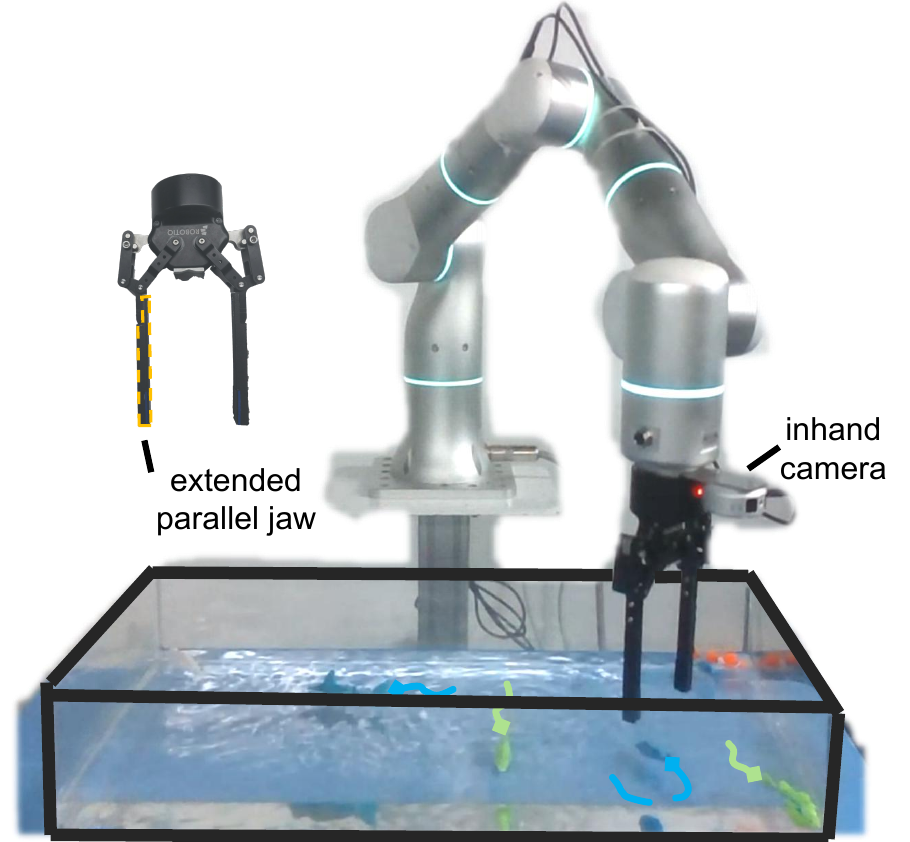}
\caption{Illustration of the hardware setting in the fish-catching task. An in-hand camera is attached to the robot's wrist. The moving trajectories of the robot fish are visualized with colored lines. We attach two 3D-printed parallel jaws to the Robotiq gripper.}
\label{fig_fish_task}
\end{figure}

For dynamic grasping, we use a Flexiv Rizon arm since it can update the servo targets more smoothly. An Intel RealSense L515 camera is attached to the wrist of the robot. We adopt the in-hand setting since the overhead camera might be occluded by the robot during tracking and choose the L515 since it has a larger depth range and can work robustly when the object is close to the camera. The D415 or D435 depth camera cannot generate depth information when the object distance to the camera is less than 15 cm. For safety reasons, we extend the fingertips of the Robotiq-85 gripper by mounting two extra 3D-printed parallel jaws. Fig.~\ref{fig_fish_task} illustrates the hardware setting in this experiment.

All the models in the following experiments run on a workstation with Ubuntu 20.04 system, Intel i9-10900K CPU, and Nvidia 2060 GPU. The code \red{is} written in Python.

\subsection{Experimental Procedure}
\label{sec:expproc}
For the parallel-jaw bin picking experiments, we conduct two independent trials for each experiment. The camera extrinsic parameters to the robot are obtained by Aruco marker detection and extra human measurement. In each trial, we randomly pour the objects onto the plate without too much human intervention. For each grasp attempt, our algorithm receives a single-view point cloud and predicts abundant grasp poses for the scene. For safety, we set a workspace limit. The limit includes the restriction on the plate plane (area enclosed by the blue dashed line on the plate, see Fig.~\ref{fig_fragments}) and the restriction on grasp pose orientation. Specifically, we restrict the angle between the grasp approach direction and the vertical direction to be less than 25$^{\circ}$, as we empirically found that a larger angle may result in an unreachable pose for the UR5 arm. The grasp pose without occlusion, located within the workspace and with the highest grasp score is selected. 

To obtain a trajectory without collision with the scene during grasping, we set a waypoint pose by translating the grasp pose 10 cm backward along its approach direction. The waypoint and grasp pose are sent to the UR5 robot arm through UR script command \texttt{movels()} via Ethernet and the motion planner of UR5 would move the gripper to these poses sequentially. We found that this simple strategy can avoid most collision situations. The gripper will close after the robot reaches the target pose, grasp the object, lift it, and then move it to the target bin. During the grasping procedure, the objects may scatter outside of the plane workspace. We would manually push these objects back. The operator will record whether the object is successfully moved to the target bin. Each trial of experiments is regarded as finished if the algorithm cannot estimate grasp poses for the scene ten times.

For the human bin-picking task, two volunteers are invited to conduct the grasping experiments and each is required to clear the plate twice.

\begin{algorithm}[!t]
\caption{Dynamic Grasping Algorithm}\label{alg:dynamic-grasping}
\begin{algorithmic}[1]
\State \textbf{Initialize}: $\textbf{\textit{robot}}.\textrm{move}(\mathbf{T}_\textrm{ready})$, open \textbf{\textit{gripper}}, empty \textbf{\textit{buffer}}
\State Mark $\mathbf{G}^0$ as \textit{empty}
\For{time step $t \leftarrow 1$ to $\infty$}
\State $\mathbf{T}^t_{tcp} \leftarrow \textbf{\textit{robot}}.\textrm{pose}$
\State $\mathcal{P}^t \leftarrow \textbf{\textit{camera}}.\textrm{perception}$
\State $\mathbf{G}^t \leftarrow \textrm{GetGraspPoses}(\mathcal{P}^t)$
\If{$\mathbf{G}^t$ is \textit{empty}} \Comment{\textcolor{blue}{lost tracking of objects}}
\State $\textbf{\textit{robot}}.\textrm{servo}(\mathbf{T}_\textrm{ready})$, empty \textbf{\textit{buffer}}
\State \textbf{continue}
\EndIf
\If{$\mathbf{G}^{t-1}$ is \textit{empty}}\Comment{\textcolor{blue}{for the first frame}}
\State $\mathcal{G}^t_* \leftarrow \arg\max_{\mathcal{G}^t\in\mathbf{G}^t}\textrm{Prob}(s = 1\mid \mathcal{P}^t,\mathcal{G}^t)$
\Else \Comment{\textcolor{blue}{for the following tracking frames}}
\State $\mathcal{G}^t_* \leftarrow \arg\max_{\mathcal{G}^t\in\mathbf{G}^t}s_\textrm{corres}(\mathcal{G}^{t-1}_*,\mathcal{G}^t)$
\EndIf
\State Add $\mathcal{G}^t_*$ into \textbf{\textit{buffer}} and predict $\tilde{\mathcal{G}}^t_*$ from \textbf{\textit{buffer}}
\State Calculate $\Delta R, \Delta t$ and $\Delta t_\textrm{xOy}$ between $\tilde{\mathcal{G}}^t_*$ and $\mathbf{T}^t_{tcp}$.
\If{$\Delta R \leq \delta_R$ \textbf{and} $\Delta t \leq \delta_t$ \textbf{and} $\Delta t_\textrm{xOy} \leq \delta_{t_\textrm{xOy}}$}
\State \Comment{\textcolor{blue}{grasp when gripper is close enough to the fish}}
\State $\textbf{\textit{robot}}.\textrm{move}(\tilde{\mathcal{G}}^t_*)$, $\textbf{\textit{gripper}}.\textrm{close}()$ 
\State Move \textbf{\textit{robot}} to the throw pose and open \textbf{\textit{gripper}}
\State \textbf{Re-initialize} the system. \Comment{\textcolor{blue}{See L1 for details}}
\Else \Comment{\textcolor{blue}{continue tracking with pregrasp pose}}
\State Calculate pregrasp pose $\mathbf{T}^t_p$ using $\tilde{\mathcal{G}}^t_*$ and $d_\textrm{pregrasp}$
\State $\textbf{\textit{robot}}.\textrm{servo}(\mathbf{T}^t_p)$
\EndIf
\EndFor
\end{algorithmic}
\label{alg1}
\end{algorithm}

For the dynamic fish-catching experiment, we conduct five independent trials. The camera extrinsic is obtained by measuring the customized camera holder. The pseudo-code of our dynamic grasping algorithm is illustrated in Alg. \ref{alg:dynamic-grasping}. In each fish-catching trial, we first generate \red{abundant grasp poses for the scene (Line. 6 in Alg. \ref{alg:dynamic-grasping})} and select the best grasp pose $\mathcal{G}_1^*$ for the first frame \red{(Line. 12 in Alg. \ref{alg:dynamic-grasping})} using the same procedure in the static scene. For simplicity, we \red{assume that the grasp pose has already been transformed into} the robot coordinate system using the camera extrinsic parameters. Then the robot enters the pre-grasp servoing stage. The grasp perception system keeps active during this process and generates updated grasp pose $\mathcal{G}^t_*$ of time step $t$ \red{(Line. 14 in Alg. \ref{alg:dynamic-grasping})} that has a high association score with the previous grasp pose $\mathcal{G}^{t-1}_*$. Empirically, we found that it is important to predict the future grasp pose when catching the moving fish since it takes some time to close the gripper. Hence, the selected grasp poses during the tracking procedure are saved to a temporary buffer of length 10 for future grasp pose prediction. The system will predict a future grasp pose $\tilde{\mathcal{G}}^t_*$ \red{(Line. 16 in Alg. \ref{alg:dynamic-grasping})} by adding the moving momentum calculated from the temporary buffer to the last grasp pose $\mathcal{G}^t_*$.  In the pre-grasp servoing stage, we do not directly servo to the future grasp pose. Instead, the servoing target pose $\mathbf{T}^t_p$ during the pre-grasp stage is obtained by translating $\tilde{\mathcal{G}}^t_*$  along the $z$-axis by $d_\textrm{pregrasp} = 3.5\textrm{cm}$ backward \red{(Line. 24 in Alg. \ref{alg:dynamic-grasping})}. This is to avoid touching the moving fish and changing their motion status during servoing. Meanwhile, we only preserve the rotation along the  $z$-axis of the gripper and eliminate the rotation along the other two axes to improve the tracking stability. The servoing target $\mathbf{T}^t_p$ is sent to the Flexiv arm through Flexiv RDK and the robot would servo to the target pose \red{(Line. 25 in Alg. \ref{alg:dynamic-grasping})}. The ending criterion for the servoing process is that the 3D distance $\Delta t$ between the predicted future grasp pose $\tilde{\mathcal{G}}^t_*$ and the robot's current end-effector pose is less than $\delta_t = 5.5\textrm{cm}$, the 2D distance $\Delta t_\textrm{xOy}$ in the horizontal plane is less than $\delta_{ t_\textrm{xOy}}=2\textrm{cm}$, and the angle $\Delta R$ between two poses is less than $\delta_R = 20^\circ$ \red{(Line. 18 in Alg. \ref{alg:dynamic-grasping})}. The robot will then enter the grasping stage which moves to the grasp pose and closes its gripper \red{(Line. 19-21 in Alg. \ref{alg:dynamic-grasping})}.
If the tracked pose moves outside of the robot's workspace or camera's view, the robot will return to the initial state, empty the temporary buffer, and choose a new grasp target \red{(Line. 7 in Alg. \ref{alg:dynamic-grasping})}. 
\red{
\subsection{Evaluation Metric}
\label{sec:metric}
In the area of robotic grasping, there are two different success rates used to evaluate the performance. The first metric, which we refer to as the attempt-centric success rate, is defined as the ratio of the number of successful grasp attempts to the total number of grasp attempts. This metric is commonly used in the literature~\cite{dexnet4, mousavian20196, graspnet}, as it measures the ability of a grasping method to successfully perform grasps on a per-attempt basis.}

\red{
The second metric, which we refer to as the object-centric success rate, is defined as the ratio of the number of successfully grasped objects to the total number of objects. This metric measures the ability of a grasping method to adapt to different objects, which is less strict than the attempt-centric success rate since it does not take into account the number of grasp attempts per object. This metric has been previously adopted by~\cite{sundermeyer2021contact} (they allow two attempts per object) and was referred to as ``completion rate'' in~\cite{qin2020s4g,zhao2021regnet}.}

\red{
It is important to note that we use the attempt-centric success rate in our work. However, to provide a clear comparison with previous work that adopted the object-centric success rate, we report the results of our proposed grasping method under both metrics in Table~\ref{tab:success-rates}.
}

\begin{table}[h]
\centering
\red{
\caption{Success Rates of Different Methods on Our Real Test Set. ``Dex.'' denotes DexNet 4.0 and ``Any.'' denotes our AnyGrasp.}
\label{tab:success-rates}
\begin{tabular}{|c|c|c|c|c|c|c|}
\hline
\multicolumn{1}{|c|}{} & \multicolumn{3}{c|}{Attempt-Centric} & \multicolumn{3}{c|}{Object-Centric} \\
\multicolumn{1}{|c|}{Object} & \multicolumn{3}{c|}{Success Rate (\%)} & \multicolumn{3}{c|}{Success Rate (\%)} \\
\cline{2-7}
 & Dex. & Any. & Human & Dex. & Any. & Human \\
\hline
Hardware & 59.3 & 81.5 & 91.4 & 97.2 & 100.0 & 100.0 \\
\hline
Snack & 52.3 & 100.0 & 93.9 & 93.9 & 100.0 & 100.0 \\
\hline
Ragdoll & 87.4 & 100.0 & 96.6 & 100 & 100.0 & 100.0 \\
\hline
Toy & 72.8 & 93.1 & 91.8 & 99.6 & 99.6 & 100.0 \\
\hline
Household & 64.6 & 85.5 & 94.4 & 98.1 & 100.0 & 100.0 \\
\hline
All & 72.2 & 93.3 & 93.9 & 98.9 & 99.8 & 100.0 \\
\hline
\end{tabular}
}
\end{table}

\section{Experimental Results}
In this section, several experiments are demonstrated to evaluate the following properties of our 7-DoF grasp perception system: (i) generalization ability to different objects and sensors; (ii) accuracy on different kinds of objects compared with humans; (iii) temporal consistency for random moving objects; (iv) efficiency of the whole perception system.

\subsection{Static Scenes}
\begin{figure}[!t]
\centering
\includegraphics[width=0.5\textwidth]{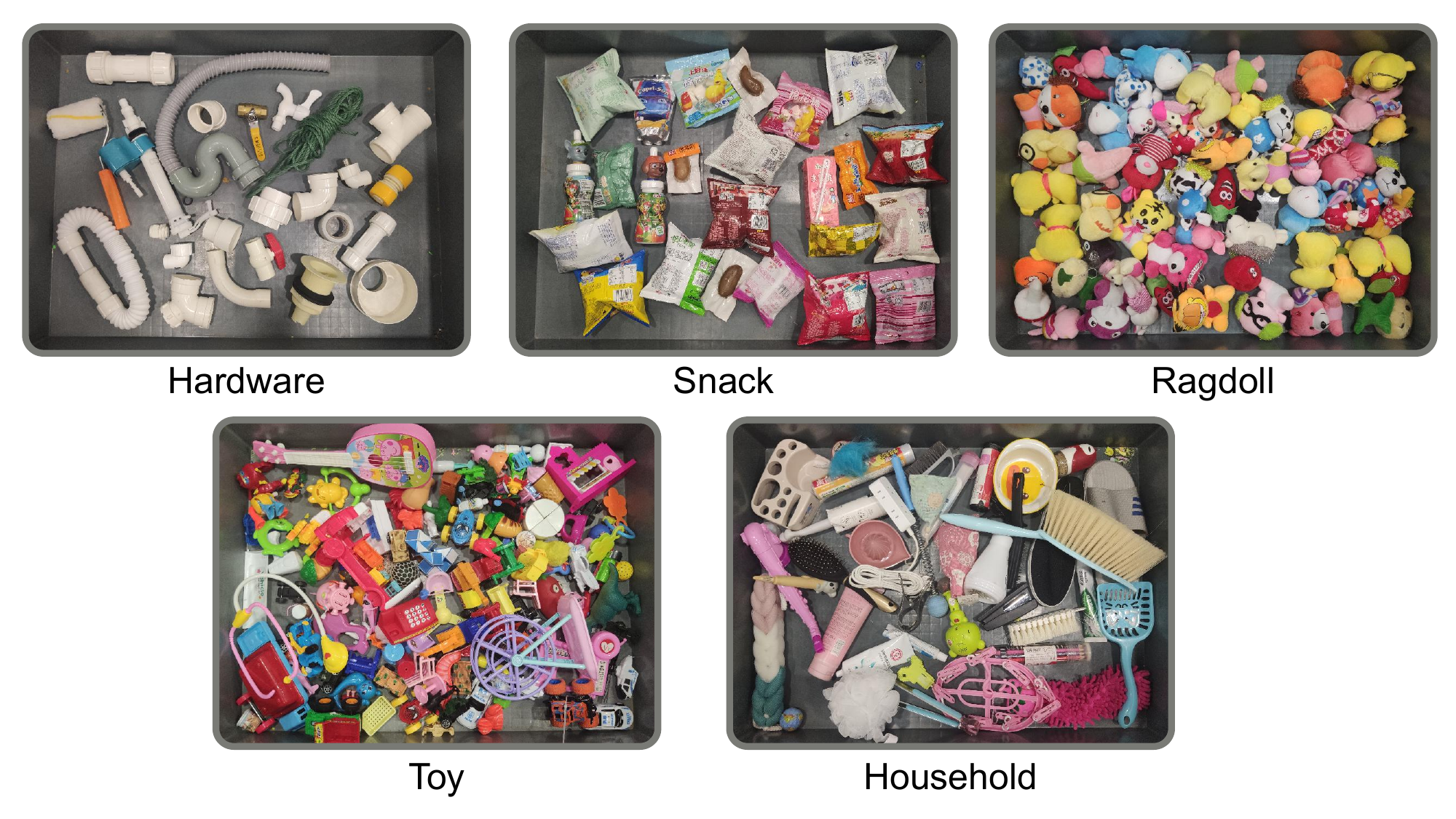}\\
\vspace{-0.1in}
(a)\\
\vspace{0.1in}
\includegraphics[width=0.5\textwidth]{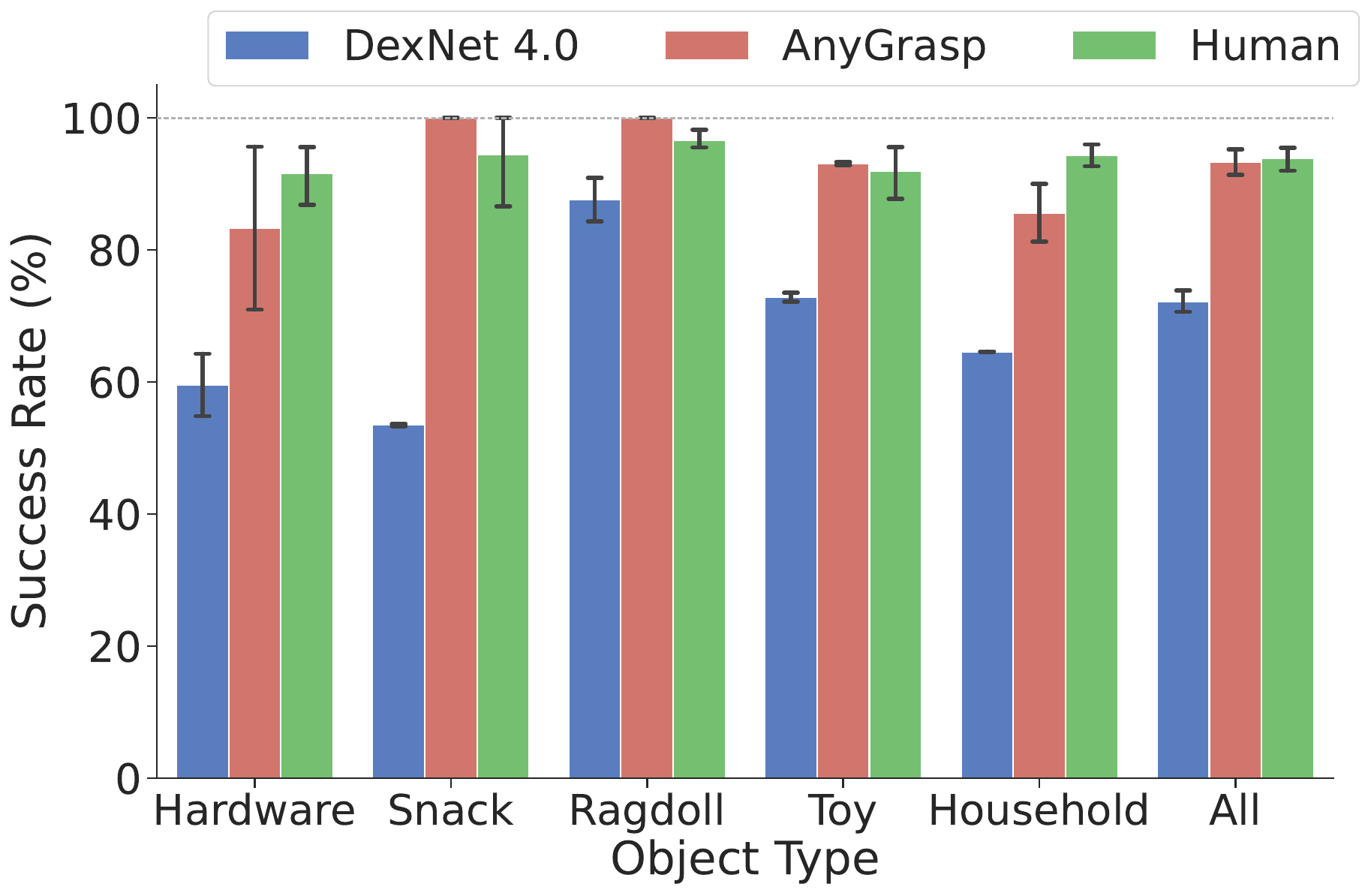}\\
(b)\\
\caption{(a) Example objects from the unseen object test set. (b) The grasp success rate of our system compared to a previous grasp pose detection system and the human's performance using a parallel jaw. }
\label{fig_obj_and_res}
\end{figure}

We first conduct grasping experiments in static scenes. To build a representative test set that can fully evaluate the grasp perception system, we collect several common categories of daily-life objects. In total, over 300 objects are collected. Fig.~\ref{fig_obj_and_res}~(a) gives an overview of the objects. 
The size of the objects ranges from $1.5 \times 1.5 \times 1.5$ cm$^3$ to $36 \times 4 \times 11.5$ cm$^3$.

We first compare the grasping system on daily objects with human operators using the same end-effector configuration and a previous method~\cite{dexnet4}. The overall grasping results on different objects are presented in Fig.~\ref{fig_obj_and_res}~(b). The detailed numerical results are given in Tab.~\ref{tab:success-rates}. The videos can be found in Appendix B (Movie S1-S4). From Fig.~\ref{fig_obj_and_res}~(b), we can see that our perception system is robust towards different categories of objects. On the large scale benchmark, its accuracy ranges from 81.5\% to 100\% on different kinds of objects, yielding 93.3\% on average. Such grasping accuracy is on-par with human subjects, which is 93.9\% on average. However, human subjects give more stable performance on different kinds of objects. The grasp accuracy ranges from 91.4\% to 96.6\%.

\begin{figure}[!t]
\centering
\includegraphics[width=0.45\textwidth]{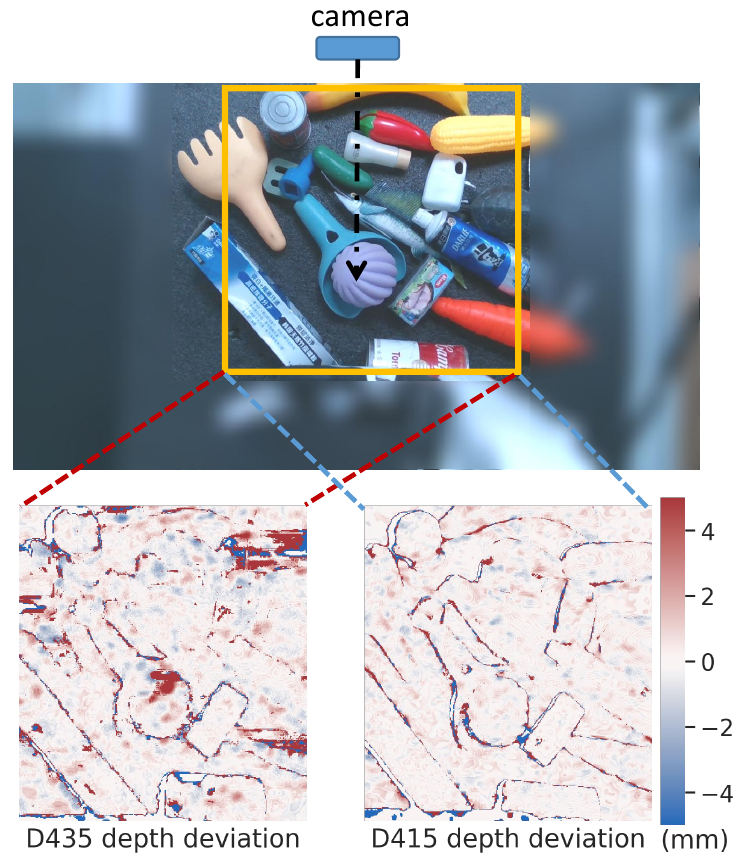}\\
\vspace{-0.1in}
(a)\\\vspace{0.1in}
\hspace{-0.3in}
\includegraphics[width=0.45\textwidth]{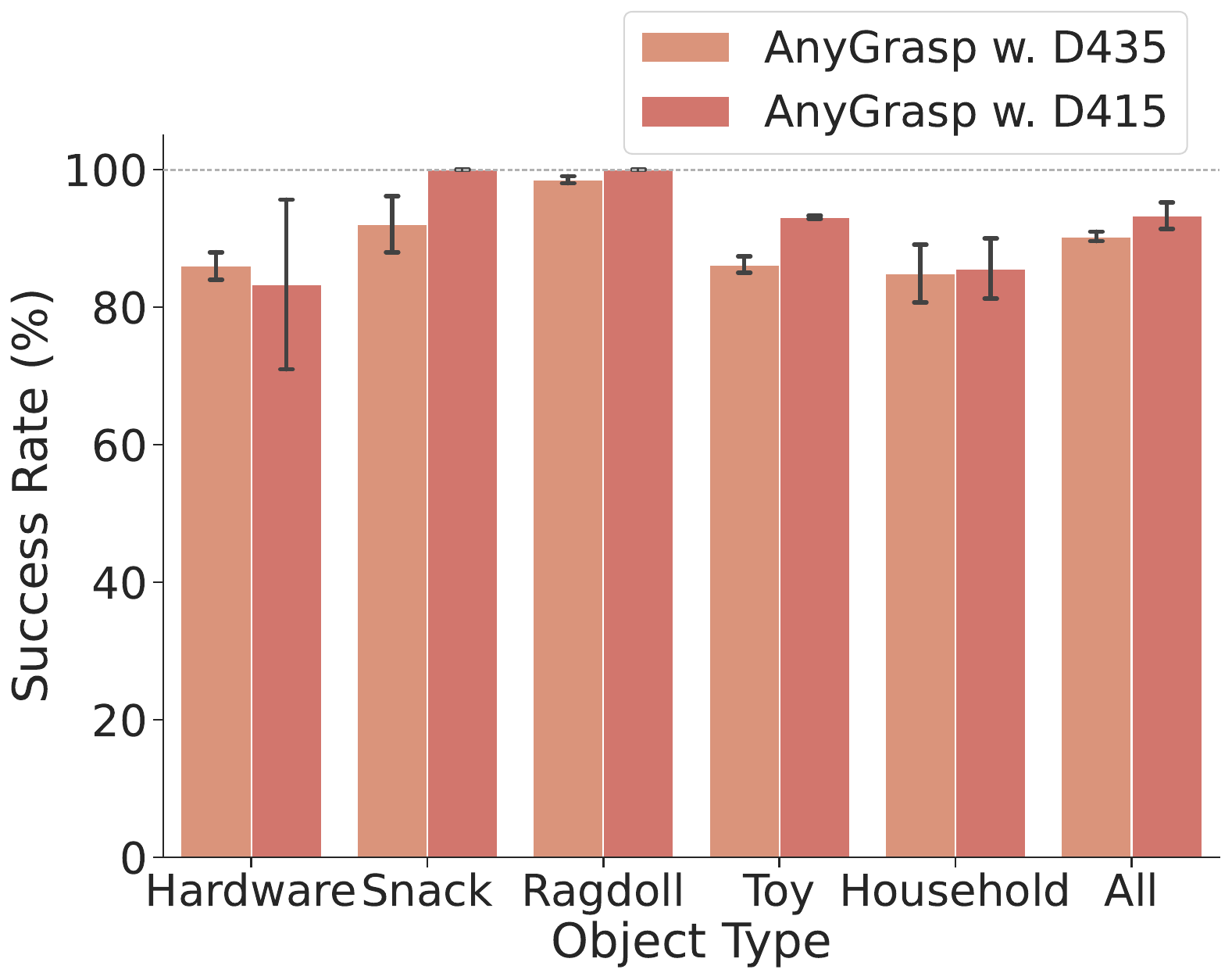}\\
(b)\\
\caption{(a) Typical depth deviation maps of two frequently adopted depth cameras. The distance unit is a millimeter. (b) The grasp success rate comparison of AnyGrasp when using these two cameras.}
\label{fig_different_camera}
\end{figure}

For each trial, the grasp perception system predicts the grasp poses in 100ms and the overall grasp decision time (including collision detection and pose adjustment) is less than 200 ms. Thus, the mean picks per hour (MPPH) only depends on the robot moving and gripper executing speed. With a single UR5 arm that has a maximum end-link speed of 1m/s and a Robotiq gripper, our grasping system achieves over 900 MPPH, which is a dramatic improvement compared to previous state-of-the-art~\cite{dexnet4}, 300 MPPH achieved by a dual-arm system. Such a metric can be further optimized in the industry using a higher-speed robot arm and gripper. We also observe that humans can achieve over 1,500 MPPH with their maximum speed, and on average they can achieve 1,000-1,200 MPPH.

We then show the performance of our algorithm on different depth sensors. We visualize the typical \red{deviation} maps of the two cameras we use in Fig.~\ref{fig_different_camera}~(a). The \red{deviation} map is obtained by subtracting \red{the mean depth value of 100 repeatedly captured images from a randomly chosen depth image.} We can see that the depth given by the D435 camera presents a larger variance and can achieve up to \red{$\pm5$}mm error. Even with this noise, our algorithm can still perform well, as illustrated in Fig.~\ref{fig_different_camera}~(b). The video of grasping using a D435 camera can be found in Appendix B (Movie S5). The experiments demonstrate the robustness of our algorithm towards depth sensing noise, mainly owing to the training data collected with real sensors.

\begin{figure}[!t]
\centering
\includegraphics[width=0.3\textwidth]{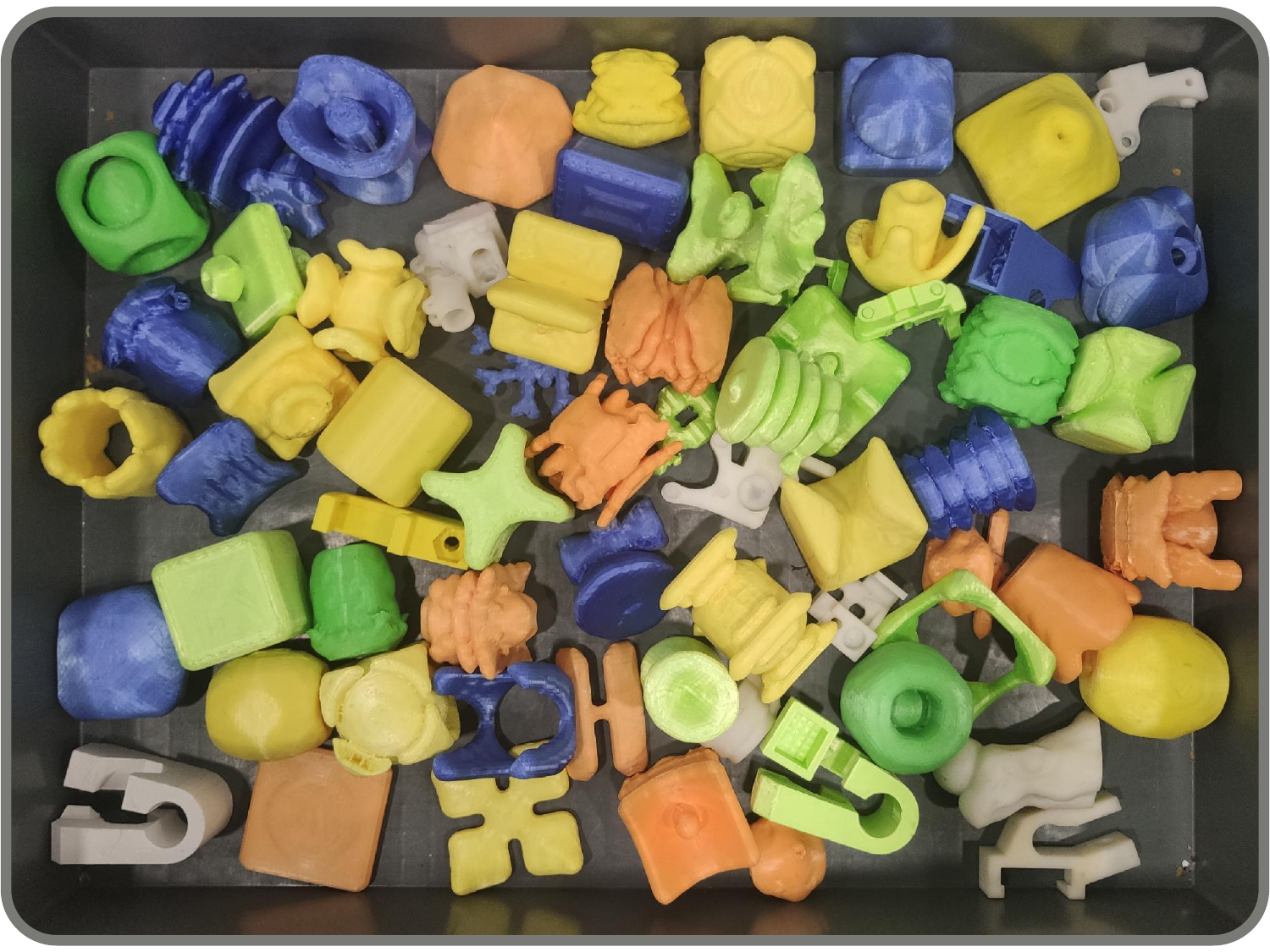}\\
(a)\\
\includegraphics[width=0.35\textwidth]{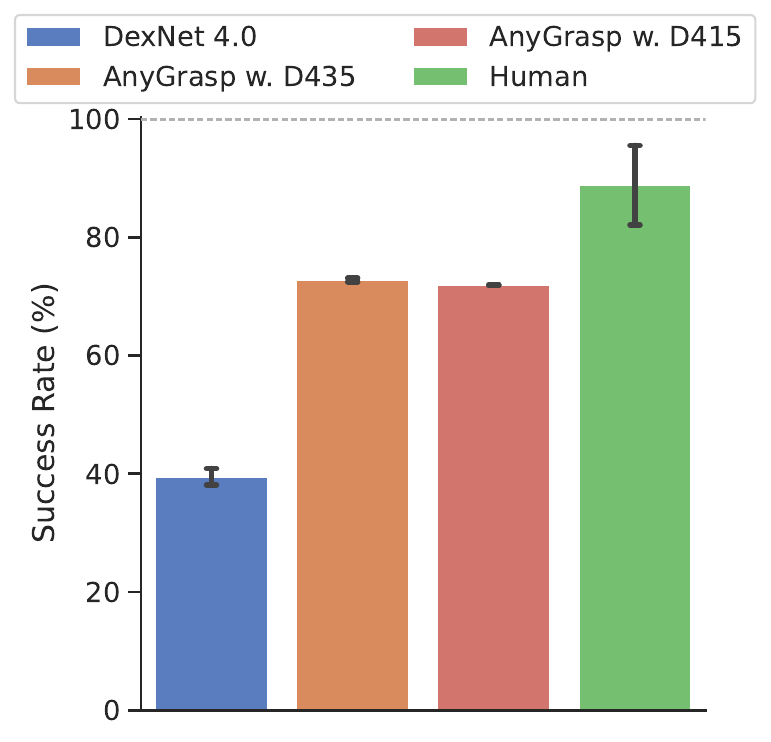}\\
(b)\\
\caption{(a) The 3D-printed adversarial objects. It can be seen that the object's surface is quite smooth. (b) The grasp success rate of different methods on the challenging adversarial object set.}
\label{fig_adversarial}
\end{figure}

\begin{figure}[!t]
\centering
\includegraphics[width=0.485\textwidth]{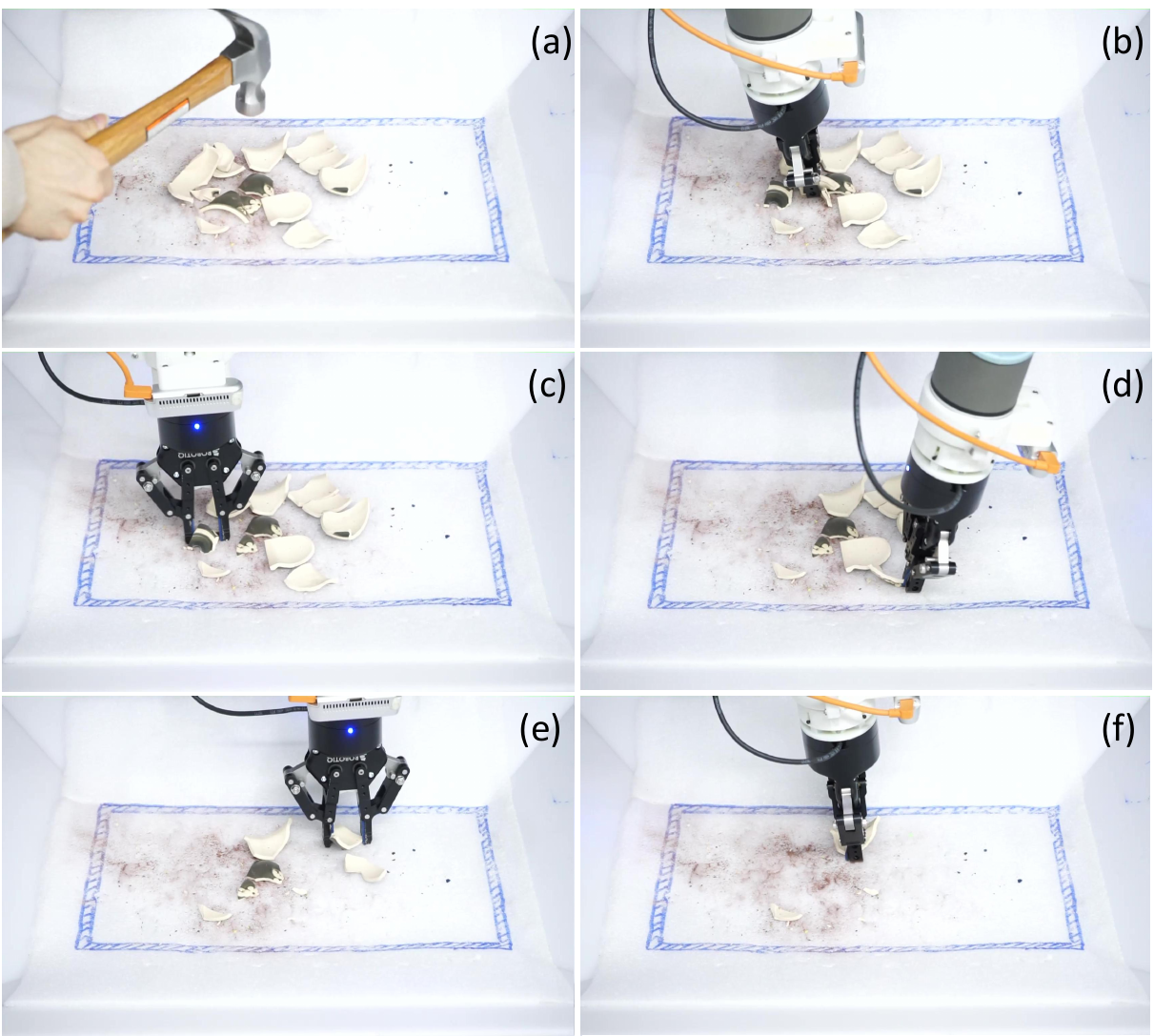}
\caption{Snapshots of the fragments cleaning task. It requires the grasp pose detection model to generate accurate poses for the thin pieces. The sequence order is from (a) to (f).}
\label{fig_fragments}
\end{figure}

We further evaluate our grasp perception system on a challenging adversarial object set, which contains the 13 adversarial objects used in DexNet2.0~\cite{dexnet2} and the 49 evaluation objects from EGAD~\cite{morrison2020egad}. The objects from the EGAD evaluation set are generated by the program to ensure the shape complexity and grasp difficulty. Fig.~\ref{fig_adversarial} shows the examples of the adversarial object set and the statistics of the success rate. Full videos can be found in Appendix B (Movie S6-S8, part of S3-S4). We can see that the accuracy decreases for our algorithm and the previous method~\cite{dexnet4}. However, humans can still perform stably. The main reason is that our grasp perception system would repeat the failed trials on an object without utilizing the feedback from each trial. It deserves more exploration in the future.

Finally, we set a challenging real-world task, where the robot is required to clean the fragments of a broken clay pot. It is challenging since the fragments are pretty thin, usually less than 3mm, which requires an accurate estimation of the grasp pose. The noisy depth perception further increases the difficulty. We haven't seen previous grasp pose detection algorithms demonstrate success in this scenario. Fig.~\ref{fig_fragments} shows the snapshots of a robot cleaning the thin fragments under the guidance of the AnyGrasp perception system and the video is given in Appendix B (Movie S9).

\begin{figure}[!t]
\centering
\includegraphics[width=0.35\textwidth]{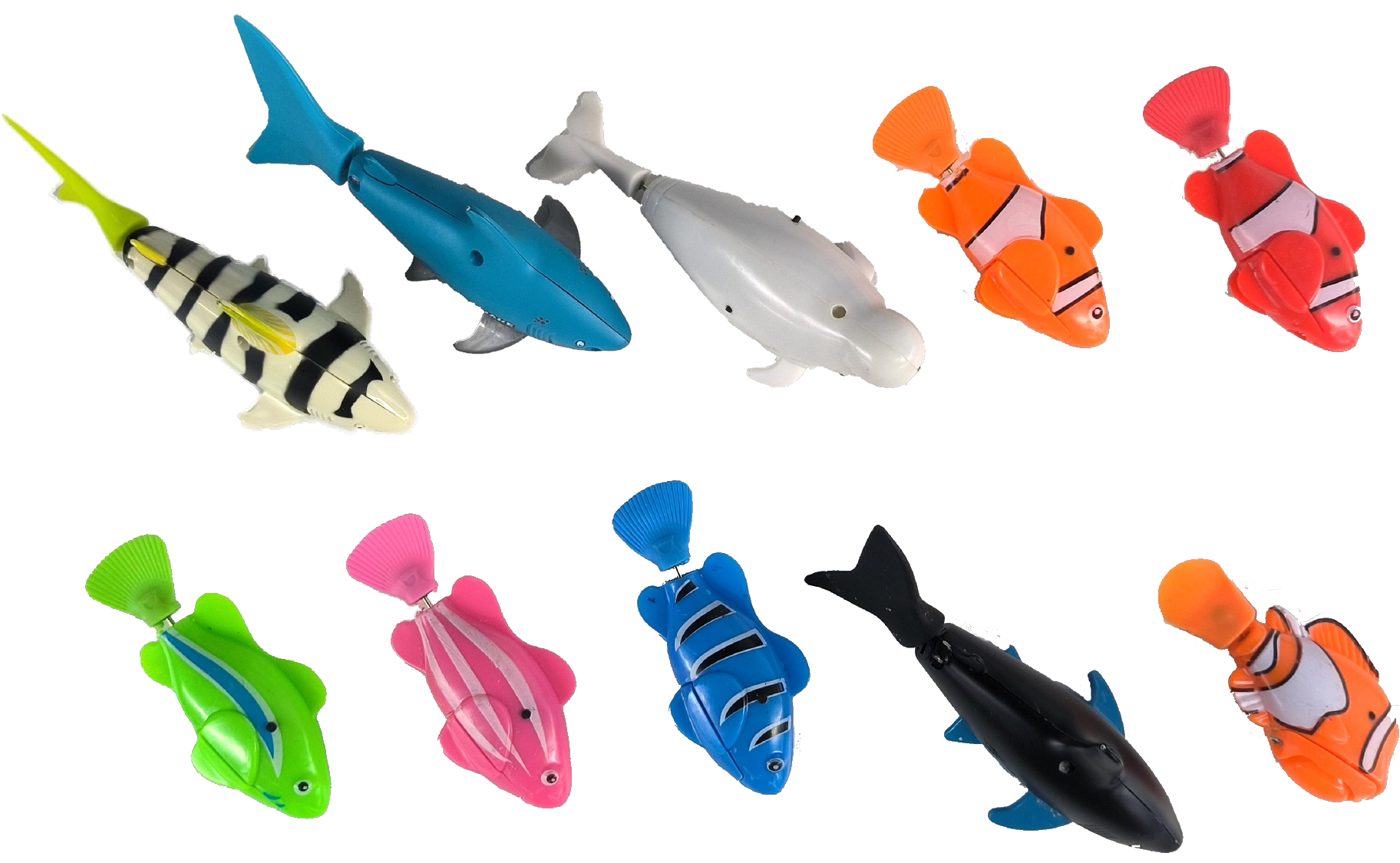}\\
(a)\\
\includegraphics[width=0.5\textwidth]{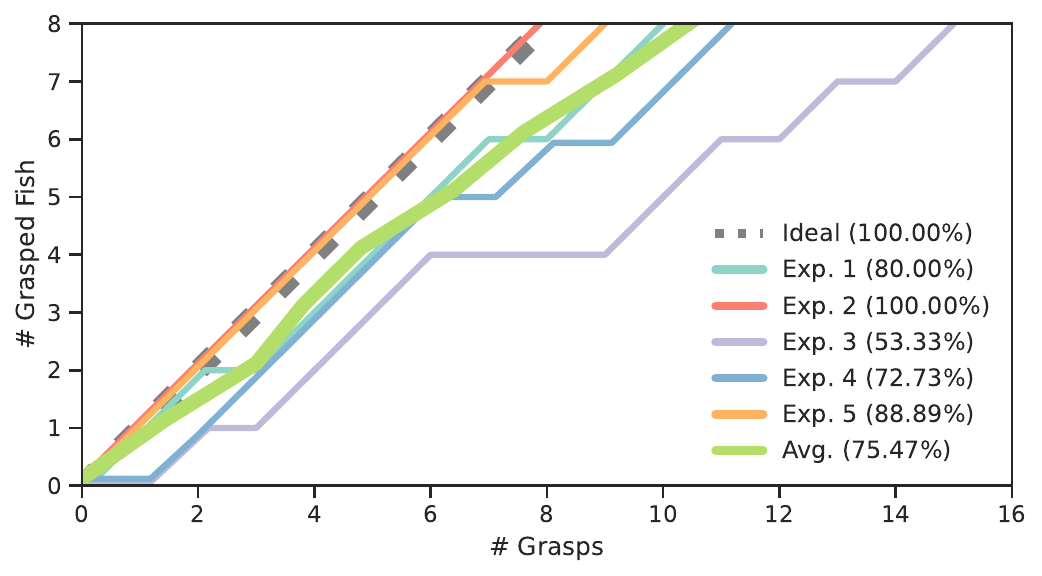}\\
(b)\\
\caption{(a) Different robot fish were used in our experiments. (b) Performance of the fish-catching experiments.}
\label{fig_fish_res}
\end{figure}

\subsection{Dynamic Scenes}
To verify the temporal consistency of the grasp perception system, we conduct a robot fish-catching experiment. A frequently used heuristic baseline~\cite{morrison2018closing} that keeps tracking the nearest grasp poses across frames is also implemented and compared with our temporal association module.

Previous literature examines the grasp pose tracking method in a human-robot hand-over setting~\cite{yang2021reactive}. For our fish-catching experiment, it poses extra challenges for the tracking system: (i) the underwater environment greatly decreases the contact friction between the gripper and the robot fish; (ii) the fish is pretty small and tolerates small final pose errors, while the handing object is usually larger; (iii) humans tend to stabilize the objects when the robot is about to grasp, while the fish keeps moving in the whole grasping process; (iv) the point cloud would be noisier due to the light reflection and refraction in water.

\begin{figure}[!t]
\centering
\includegraphics[width=0.48\textwidth]{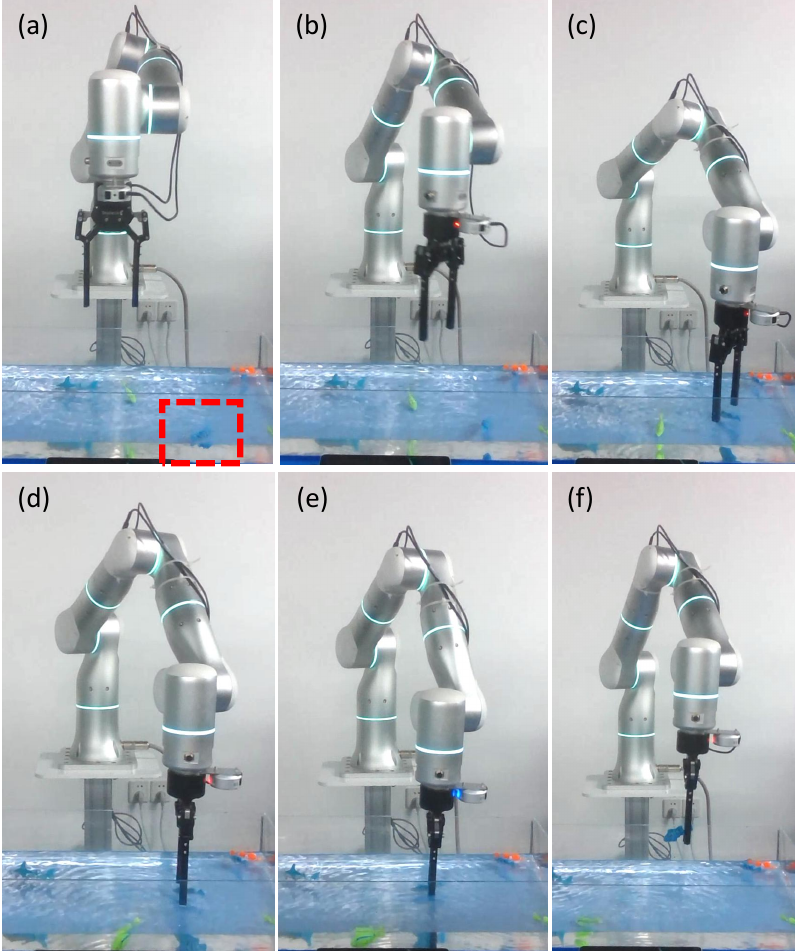}
\caption{Snapshots of the robot fish-catching procedure: (a) robot selects a grasp target, denoted by a red dashed line; (b)-(e) robot servos to the target grasp pose while updating the pose; (e) robot closes the gripper; (f) robot lifts the grasped fish.}
\label{fig_fish_demo}
\end{figure}

In our experiment, we randomly place 8 robot fish in the fish tank each time. Fig.~\ref{fig_fish_res}~(a) shows different robot fish used in our experiments. Note that we have multiple instances for each kind of fish. The robot will try to catch all the swimming fish, and we record the success rate during grasping. The whole procedure is repeated 5 times, and we show the detailed performance in Fig.~\ref{fig_fish_res}~(b). We illustrate the process of the robot catching a fish in Fig.~\ref{fig_fish_demo} and the full video can be found in Appendix B (Movie S10). The average success rate achieves 75.5\%. We can see that our grasp perception system is robust in this challenging dynamic grasping scenario. 

\begin{figure}[!t]
\centering
\includegraphics[width=0.45\textwidth]{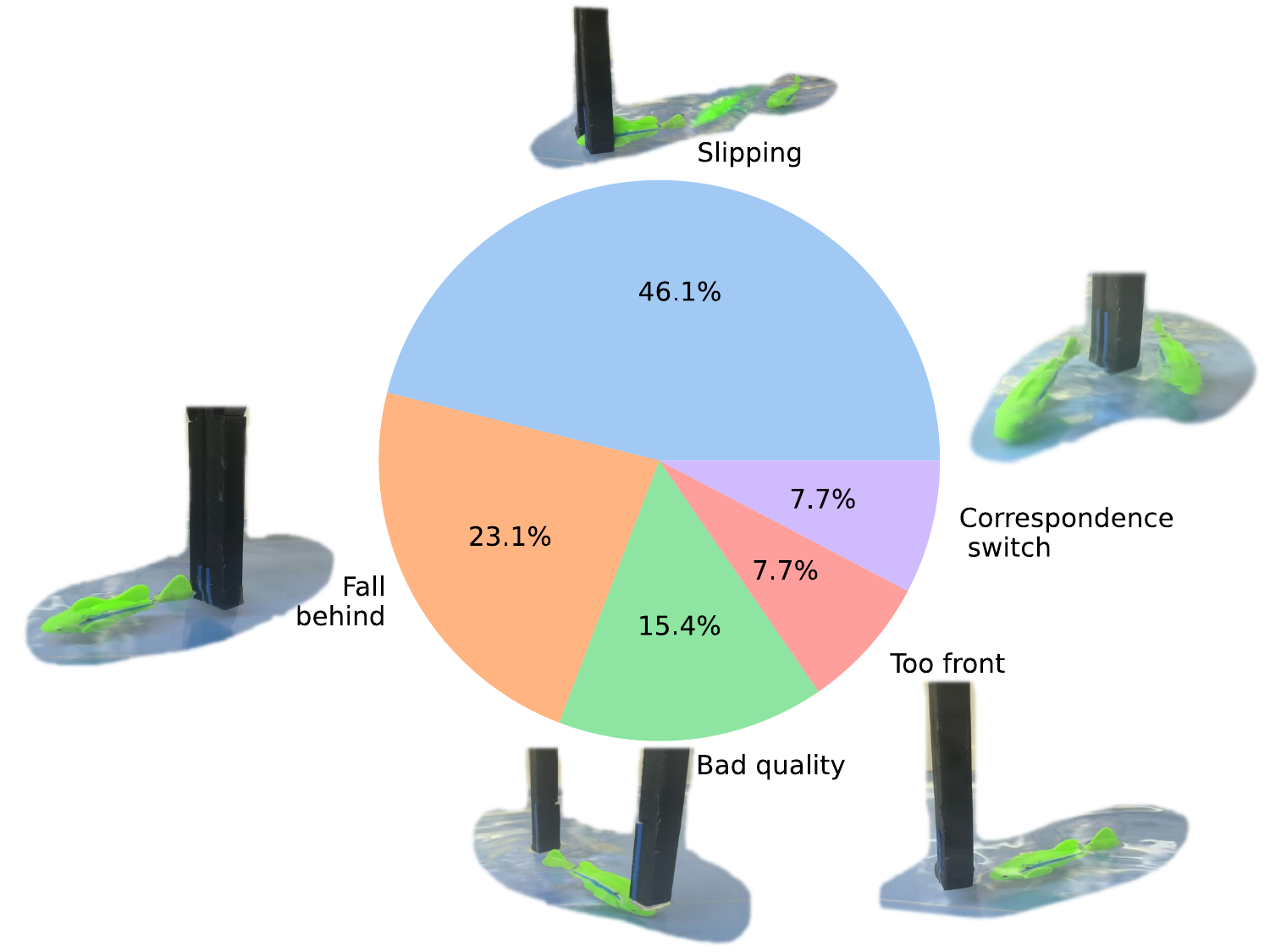}
\caption{Failure analysis in the fish catching experiments. We illustrate each failure case with an example image. Zoom in for more details and see the text for more explanation.}
\label{fig_failure}
\end{figure}

We further conduct a failure analysis and found that the failed cases can be divided into five categories, as shown in Fig.~\ref{fig_failure}. In nearly half of the failure cases, the fish slips away although the grasp pose is good, mainly due to the small friction in the water. The second and fourth reasons are that the predicted future grasp pose falls in front of or behind the fish. The main reason is that the robot fish would change its speed during moving and the historical momentum is outdated. The third reason is that the grasp quality is not good enough, and the gripper finger would push the fish away during closing. The last reason is the correspondence switch, which happens when there are two close and similar fish. The correspondence switch would inject noise into the history pose buffer and lead to a wrong predicted future grasp pose.

On the other hand, the heuristic method achieves an average success rate of 62.5\%. During grasping, we found that this nearest target policy is more likely to fall behind the moving target during tracking, and takes a longer time to enter the grasping state. Specifically, it takes 12.7\% more time than our method on the successful grasps on average. 

Overall, the grasp perception system can run at 7 Hz on an Nvidia 2060 GPU.

\section{Discussion}

\subsection{Train in Simulation}
As most methods and datasets provide training data in simulation, we try to figure out the performance gap between training on real data and simulated data. Since our dataset contains object 3D models and their 6D poses in the scene, we obtain simulated RGBD images by rendering the scene in PyRender~\cite{matl2019pyrender}. The same model is trained on these simulated depth images. We also try the frequently adopted data augmentation technique that adds Gaussian noise to the depth images~\cite{dexnet2, dexnet4} for better sim-to-real transfer.

\begin{figure}[!t]
\centering
\subfloat[]{\includegraphics[width=0.22\textwidth]{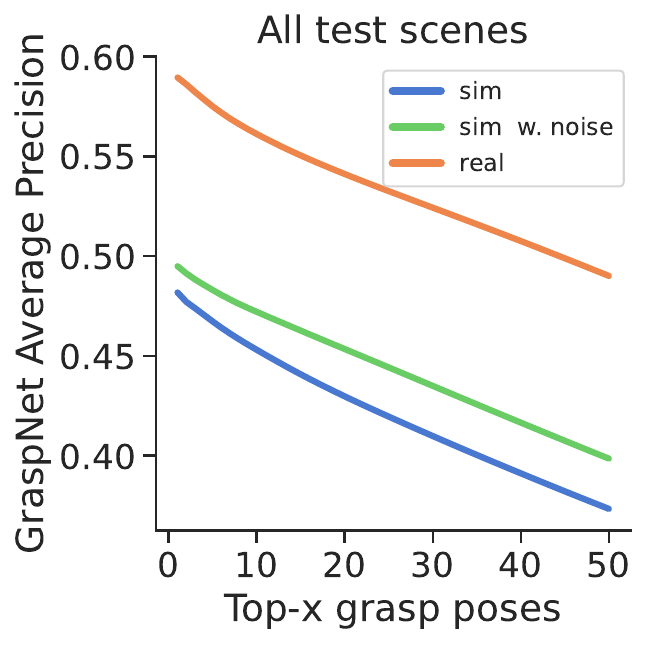}}
\hfil
\subfloat[]{\includegraphics[width=0.22\textwidth]{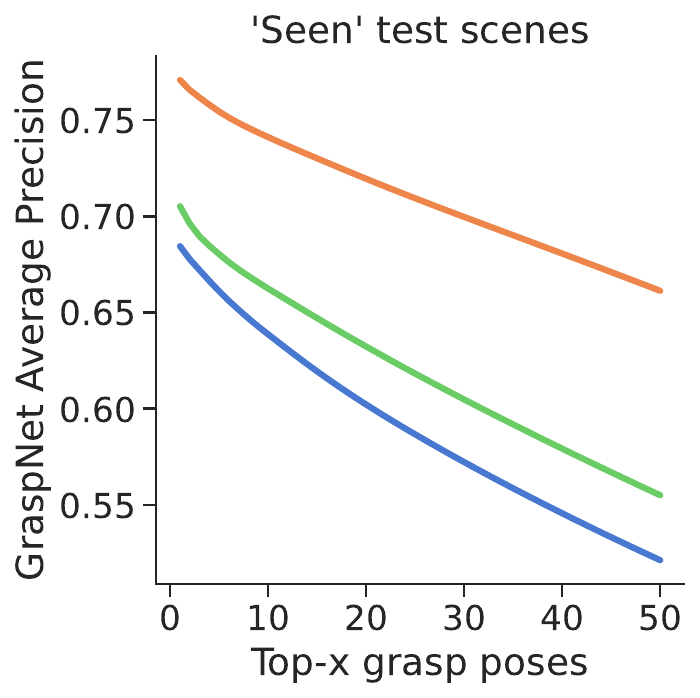}}
\hfil
\subfloat[]{\includegraphics[width=0.22\textwidth]{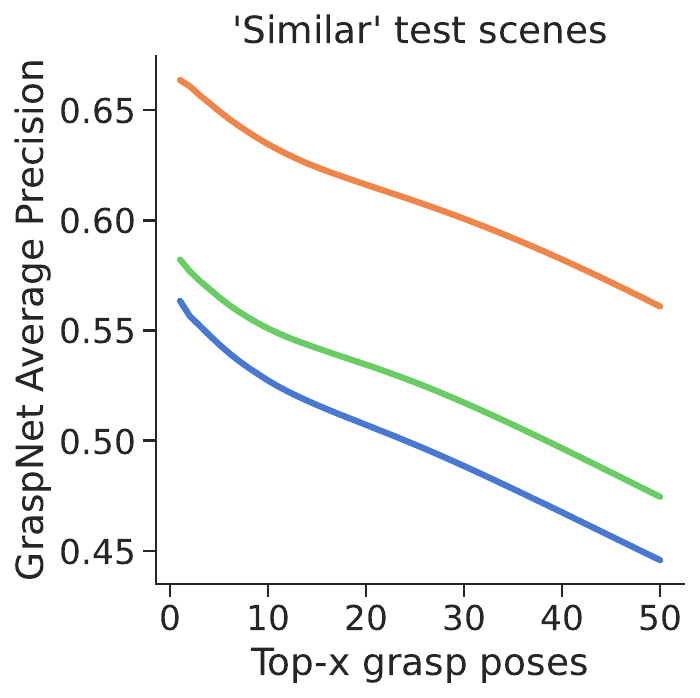}}
\hfil
\subfloat[]{\includegraphics[width=0.22\textwidth]{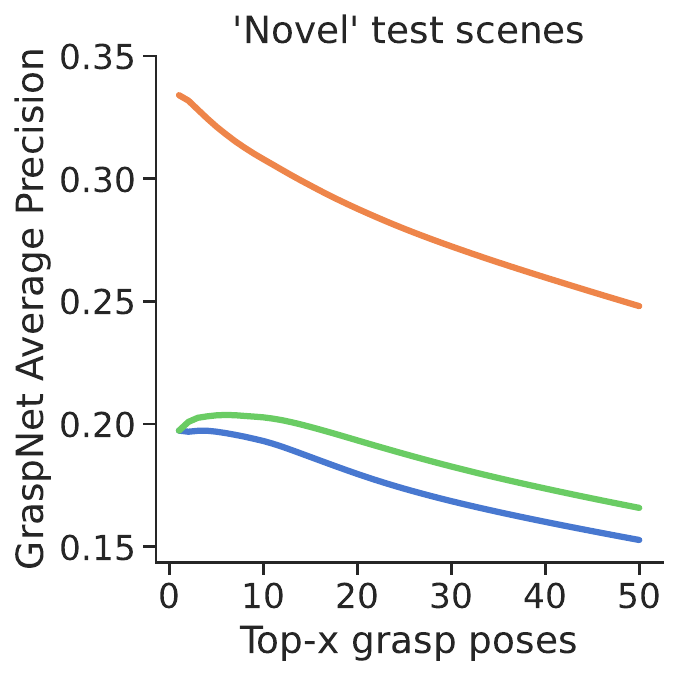}}
\caption{Evaluation results on the GraspNet-1Billion test set when training with simulated and real-world data. From (a) to (d) we present the average precision (AP) on the whole test set, the test set with seen objects in the training set, the test set with similar objects in the training set, and the test set with novel objects. ``sim'' denotes the model trained with perfect simulated depth images, ``sim w. noise'' denotes the model trained with simulated depth with Gaussian noise.}
\label{fig_sim_res}
\end{figure}

We first analyze the detailed performance on the GraspNet-1Billion benchmark, which directly evaluates the grasp pose's quality based on object mesh using the force-closure analysis. The original metric computes the averaged AP for the Top-50 predicted grasp poses in the scene. To show more details, we compute the averaged AP for grasp poses ranging from Top-1 to Top-50. The performance on different test splits for our network trained with different strategies is shown in Fig.~\ref{fig_sim_res}. We can see that adding Gaussian noise to the point cloud can improve the performance during \red{evaluation}, but still remains a large performance gap with directly trained on real data. Besides, their performance gap widens when the test set contains harder objects (the novel test scenes), especially for the high-score grasp poses.

Then we evaluate the model trained on simulated data with Gaussian noise in the real-world bin-picking experiment. The results are shown in Fig.~\ref{fig_sim_robot}. Movie S11 in Appendix B records the grasping process. We can see that the performance decreases a lot compared to the original model. Besides the lower success rate, we also observe that the network cannot generate grasp poses for a scene when there remain few objects. We conjecture the model trained in the simulation would generate fewer grasp poses with high scores in the real world due to the domain shift.

\red{In this experiment, we showed that the frequently adopted sim-to-real technology in the grasping community is insufficient. We encourage future researchers to explore different sim-to-real techniques (\eg~\cite{moosmann2022transfer}) and hope~\cite{graspnet} can be used as a benchmark for this direction.}

\begin{figure}[!t]
\centering
\includegraphics[width=0.45\textwidth]{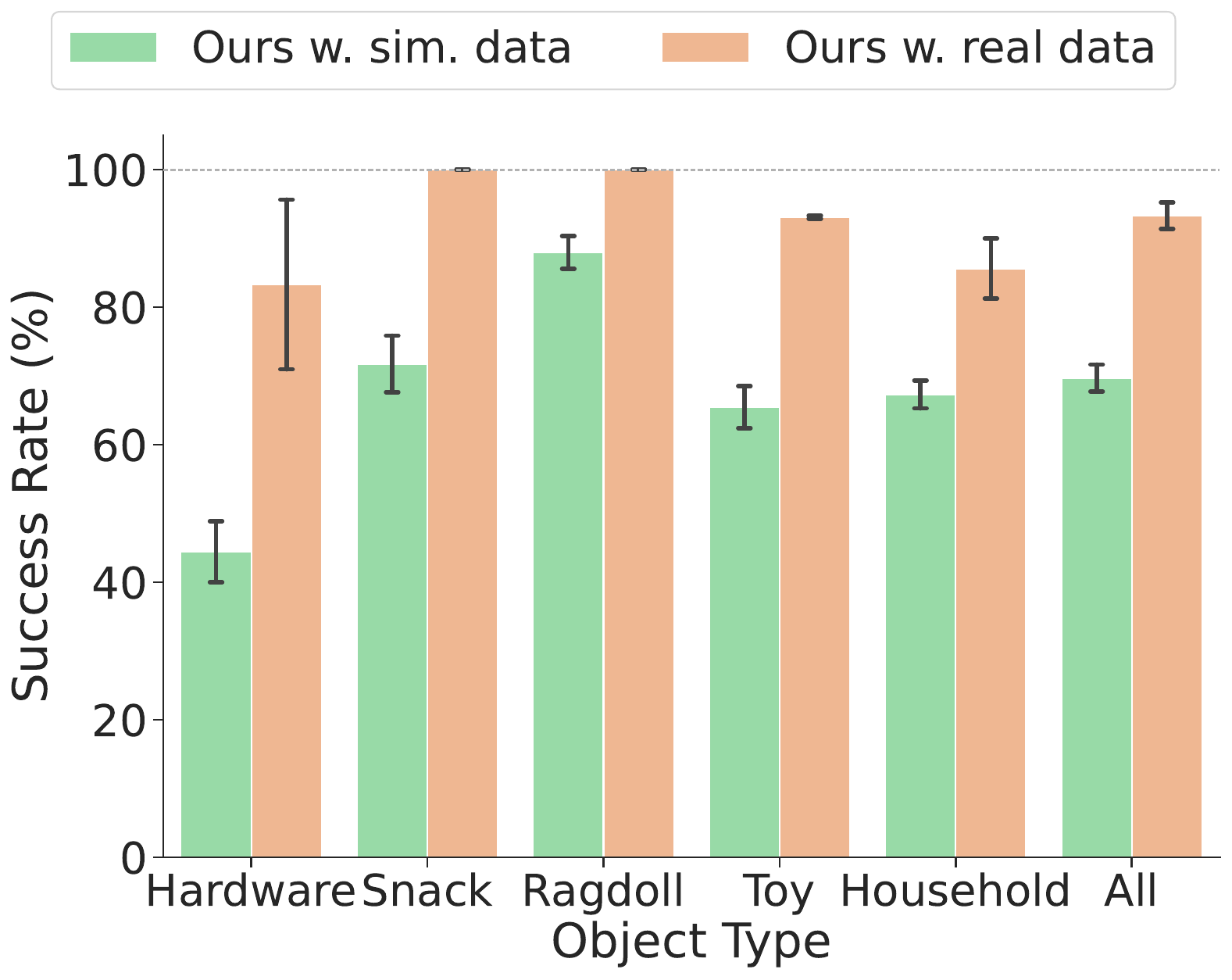}
\caption{The success rate of our system when training with real and simulated data. Note that here the ``sim. data'' denotes the simulated depth images with extra Gaussian noise.}
\label{fig_sim_robot}
\end{figure}

\begin{figure}[!t]
\centering
\subfloat[]{\includegraphics[width=0.12\textwidth]{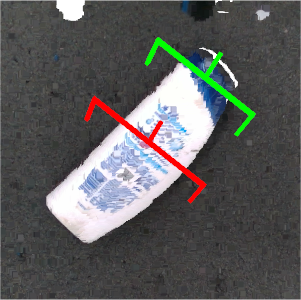}}
\hfil
\subfloat[]{\includegraphics[width=0.12\textwidth]{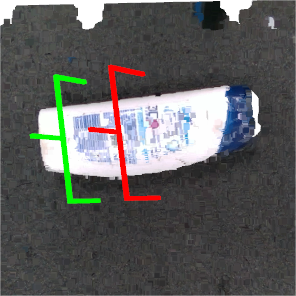}}
\hfil
\subfloat[]{\includegraphics[width=0.12\textwidth]{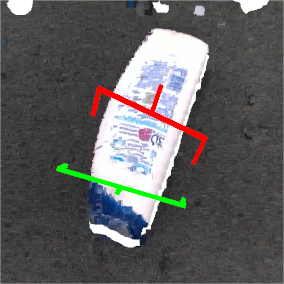}}
\hfil
\subfloat[]{\includegraphics[width=0.12\textwidth]{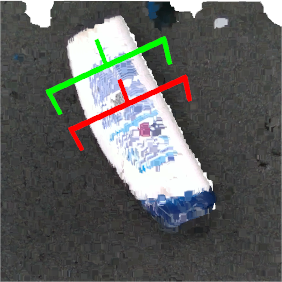}}
\caption{\red{Visualization of the top-ranking grasp poses with and without considering the stable score. The red grasp pose represents the case where we considered the stable score, while the green pose is the case without it. We see that the red grasp pose is closer to the COG of the object.}}
\label{fig_cogvis}
\end{figure}

\red{\subsection{Influence of the Stable Score}
To investigate how awareness of an object's center of mass can benefit stable grasping, we conducted an analytical experiment. We selected a long, heavy object from our test set as it better reflects the importance of COG awareness during grasping. We conducted grasping for the object with and without taking the stable score into account separately. In each experiment, we made 25 attempts, and we counted how many times in-hand slippage occurred. In total, 16 in-hand slippages (including 3 failed grasps) occurred when the stable score was not considered, and 11 in-hand slippages (including 2 failed grasps) occurred when the stable score was considered. Figure~\ref{fig_cogvis} illustrates some examples of the highest-ranking grasp poses with and without considering the stable score.}

\begin{figure*}[!t]
\centering
\subfloat[]{\includegraphics[width=0.22\textwidth]{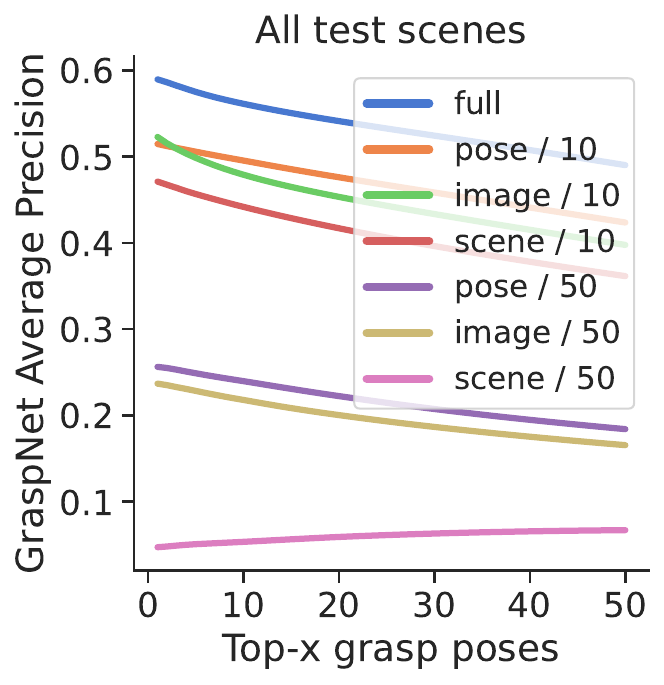}}
\hfil
\subfloat[]{\includegraphics[width=0.22\textwidth]{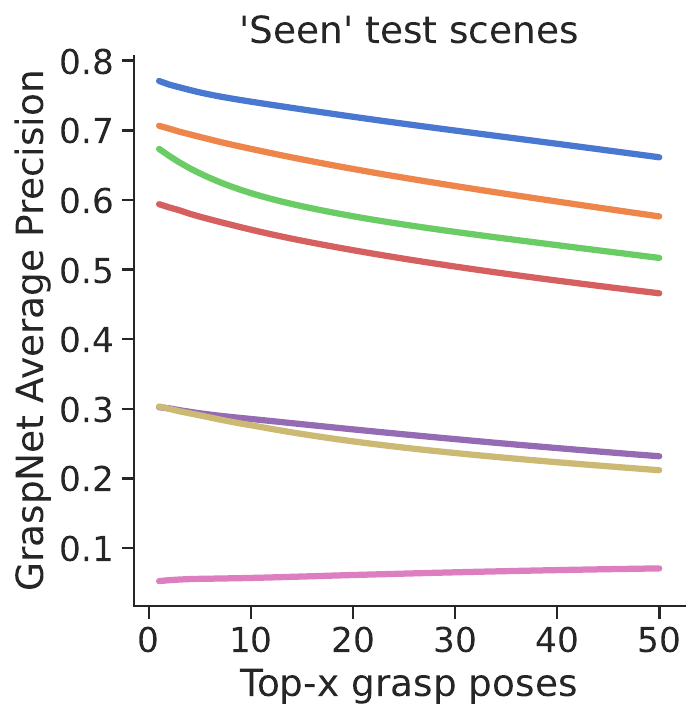}}
\hfil
\subfloat[]{\includegraphics[width=0.22\textwidth]{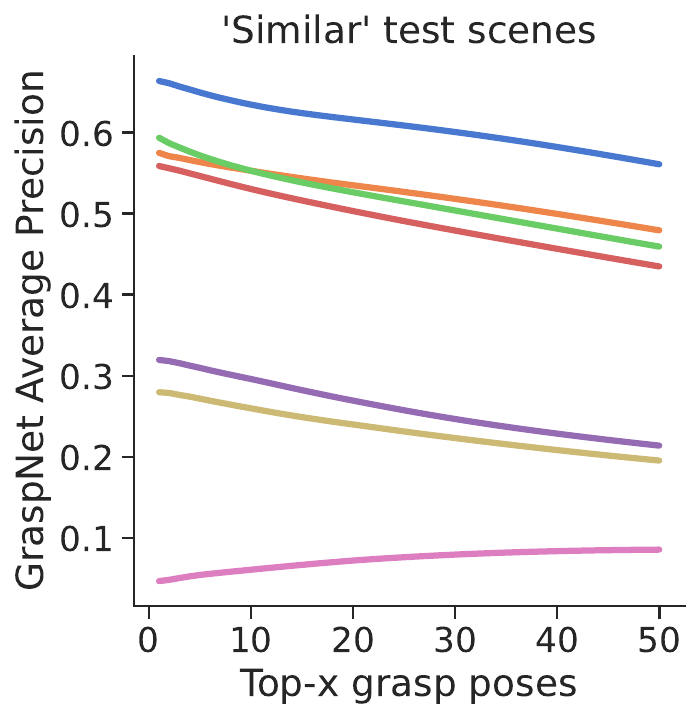}}
\hfil
\subfloat[]{\includegraphics[width=0.22\textwidth]{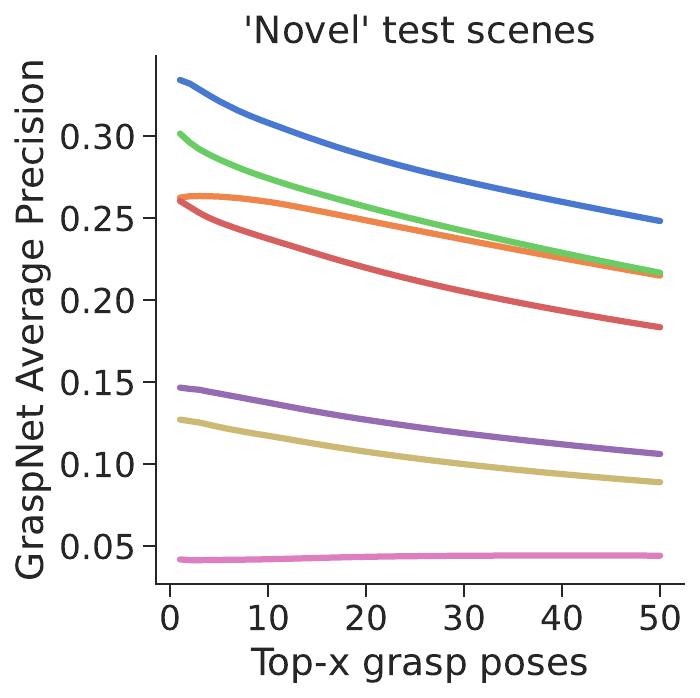}}
\caption{Evaluation results on the GraspNet-1Billion test set when training with different portions of real-world data. ``pose / 10'' denotes we downsample the grasp pose density on each object by 1/10. ``image / 10'' denotes that we downsample the training images by 1/10. `` scene / 10'' denotes that we downsample the training scene by 1/10. So are the meaning of ``pose / 50'', ``image / 50'' and ``scene / 50''.}
\label{fig_different_data}
\end{figure*}

\begin{figure*}[!t]
\centering
\includegraphics[width=0.975\textwidth]{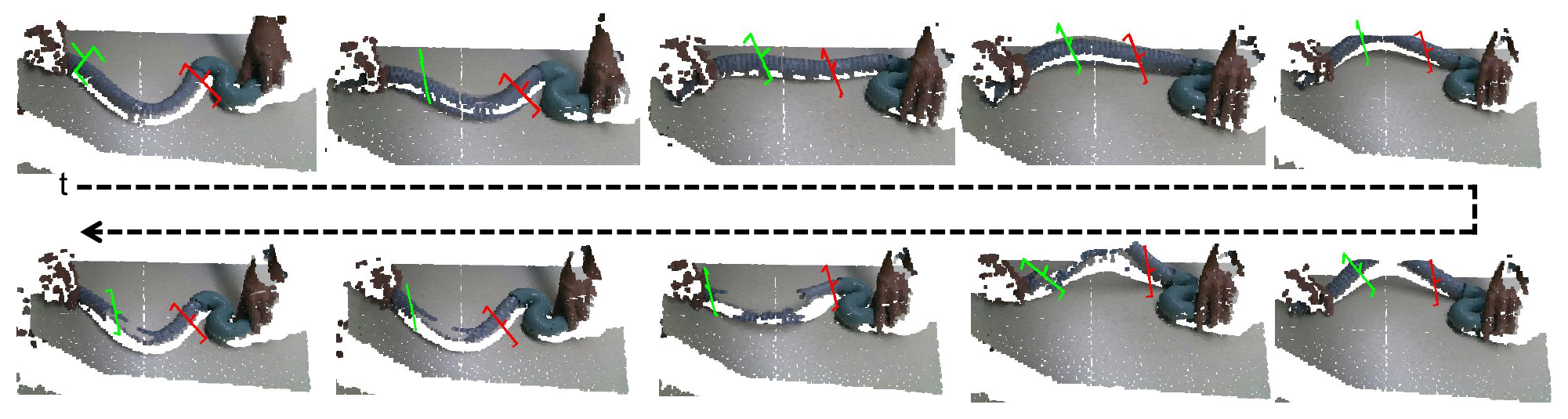}
\caption{Illustration of tracking two sampled grasp poses on a deformable, textureless tube. We can see that the grasp poses are stable in this case, even under severe noise input. The object 6D pose tracking algorithm cannot handle this scenario. }
\label{fig_tracking}
\end{figure*}

\subsection{Dense Supervision Strategy}
Besides providing data captured in real-world, another difference between GraspNet-1Billion~\cite{graspnet} and other datasets~\cite{depierre2018jacquard,dexnet2,morrison2020egad, eppner2021acronym} is that the grasp pose annotation is much denser in~\cite{graspnet}. For examples, Jacquard~\cite{depierre2018jacquard}, DexNet 2.0~\cite{dexnet2} and EGAD~\cite{morrison2020egad} generates 100 grasp poses on each object, ACRONYM~\cite{eppner2021acronym} generates 2k grasp poses and \cite{mousavian20196} generates 34k grasp poses. In contrast, the GraspNet-1Billion generates over 10M grasp poses on each object. Here we also evaluate the influence of this factor. We train multiple models on the dataset with different numbers of annotations. The models are evaluated on the GraspNet-1Billion test set with the same metric introduced in the above subsection.

In Fig.~\ref{fig_different_data}, we show the model evaluation results with different training configurations. We downsample the training set at different dimensions, namely the grasp pose density on each object, the image amount, and the scene amount. These dimensions are uniformly downsampled by 10 times and 50 times respectively. We can see that when downsampling the grasp poses on objects by 10 times, the performance degradation is similar to decreasing the number of training images. Meanwhile, the performance decreases more \red{in} the novel object test scenes. When we downsample the training samples by 50 times, we can see that the degradation is still similar for the grasp pose dimension and the image dimension. These results suggest that the densely annotated grasp poses are as equal importance as the number of training images. Considering that they can be annotated automatically without extra human efforts in collecting images, it is a good trade to annotate dense grasp poses on objects. 

Another interesting finding is that when we downsample the scene dimension of the training data, the model performance shows larger degradation on the test sets. When we downsample the scene by 50 times (namely 2 training scenes are used), the model cannot converge. It suggests that the diverse scene might be more important for model training. We hope our analyses would be beneficial to future dataset design.

\subsection{Comparison with 6D Pose Tracking}
Recently, there is also research focusing on novel object 6D pose tracking~\cite{wen2021bundletrack}. Thus, it may be intuitive to adopt an unseen object 6D pose tracking pipeline for grasp tracking. Here, we explain the advantages of grasp pose tracking versus object 6D pose tracking. Firstly, the 6D pose tracking algorithm cannot handle deformable objects, and thus is not applicable for grasping in such cases. Secondly, the 6D pose tracking algorithm may be prone to large depth noise like incomplete point cloud, since it focuses on the object level features. But the grasp tracking algorithm can work robustly for a target grasp pose, even though other parts of the object are heavily occluded. In Fig.~\ref{fig_tracking} we show an example illustrating these two cases. 

\subsection{Closed-loop Grasp Adjustment under Occlusion}
Humans conduct closed-loop grasping in daily manipulation. Although our algorithm is also temporal continuous and can enable closed-loop grasping, it is fragile to the occlusion induced by the robot. When the robot approaches the object, the gripper would occlude the grasp target. Thus, the visual perception would fail. How to enable the closed-loop grasp perception against occlusion is an open question. A missing piece here is tactile perception. It aids grasp slip detection and enables in-hand adjustment. This ability is out of the scope of this paper and will be our future work.

\section{Conclusion}
In this paper, we presented AnyGrasp, a visual grasp perception system that can generate spatially dense and temporally smooth grasp poses. We proposed a unified model, wherein the dense prediction of the geometry processing module is the basis to enable smooth grasp pose tracking in the temporal association module. The collisions and object COG were implicitly learned through the supervision signal. 
Owing to our learning strategy with real-world data, the model performs robustly against different real-world depth sensor noise. Various experiments were conducted to verify the accuracy, robustness, and efficiency of our method. Finally, we analyzed the influence of several factors in the dataset design.

Although our method provides the closed-loop grasping ability, it cannot adjust the grasp pose using visual or tactile feedback. This ability can help the robot to recover from previous errors soon. Meanwhile, we only focused on visual grasping for the two-finger parallel gripper in this section. It would be significant to transfer such ability to different robotic hands. Our future research will be conducted accordingly.

\section*{Acknowledgments}
\small{This work is supported in part by the National Key R\&D Program of China, No. 2017YFA0700800, Shanghai Municipal Science and Technology Major Project (2021SHZDZX0102), Shanghai Qi Zhi Institute, SHEITC (2018-RGZN-02046). We would like to thank Jiang Zou and Xiaolin Fang for their insightful discussions and comments.}

\small{\textit{Authors' contributions: }H.-S.F. designed the experiments, devised the training method, implemented the static grasping experiments, collected the data and wrote the manuscript. C.W. implemented the neural network training, annotated the grasp label and edited the manuscript. H.F. implemented and conducted the dynamic grasping experiments. M.G. helped the data collection, wrote the API for data parsing and built the robot platform. J.L. generated the simulated data and assisted real-robot experiments. H.Y and W.L assisted real-robot experiments. Y.X. assisted the network training. C.L. supervised the project and provided the environment and funding support for this research.}

{\appendix[]

\subsection*{A. Object Collection}
To comprehensively evaluate the algorithm performance in daily scenario, we decide to collect a fairly large scale test object set, as we found that the results varied on different objects. To construct the object set, we go to the supermarket, the hardware store and the toy store for procurement. The principle for selecting objects is that there exists some graspable place smaller than the gripper width on a object. In total, over 300 objects are collected, which is an order of magnitude larger than previous practice~\cite{dexnet4}. We believe such a large object test can provide a thorough evaluation to algorithms. Note that we do not collect objects with a large portion of transparent or black surface, since current depth sensor cannot give good prediction on these materials. Some recent papers~\cite{sajjan2020clear,fang2022transcg} focused on solving this problem with RGB information, but it is not fully addressed yet.The challenging adversarial object set is 3D printed with poly-lactic acid material. This results a smooth surface and makes the grasping even harder.
\subsection*{B. Supplementary Videos}
We record all the real robot experiments conducted in this paper. All the videos are uploaded to YouTube and will be permanently stored in support of this paper. The original length of the videos is over 12 hours. They are speeded up according to their importance to the paper. We honestly report the statistics of each video. Below we provide the links to these supplementary videos.
\begin{itemize}
    \item S1: ``AnyGrasp Demo: RealSense D415 camera on daily objects'', \url{https://youtu.be/dNnLgAGreec} 
    \item S2: ``DexNet4.0 tested on daily objects'', \url{https://youtu.be/vDqsrj_rtk8}
    \item S3: ``AnyGrasp experiment: Human subject 1 on daily objects and adversarial objects'', \url{https://youtu.be/-h0yvFZDfko}
    \item S4: ``AnyGrasp experiment: Human subject 2 on daily objects and adversarial objects'', \url{https://youtu.be/yItVN-Awjrg}
    \item S5: ``AnyGrasp Demo: RealSense D435 camera on daily objects'', \url{https://youtu.be/7pgdbyLN0A4}
    \item S6: ``DexNet4.0 tested on adversarial objects'', \url{https://youtu.be/9vOop28YReg}
    \item S7: ``AnyGrasp Demo: RealSense D435 camera on adversarial objects'', \url{https://youtu.be/AK_nHgH4RBA}
    \item S8: ``AnyGrasp Demo: RealSense D415 camera on adversarial objects'', \url{https://youtu.be/8FztVFRcvMY}
    \item S9: ``AnyGrasp Demo: Cleaning fragments of a broken pot'', \url{https://youtu.be/s0SUw1vgtr8}
    \item S10: ``AnyGrasp Demo: Robot Fish Catching by A Robot'', \url{https://youtu.be/2O7UoOxeLlk}
    \item S11: ``AnyGrasp trained solely in simulation, tested in real world'', \url{https://youtu.be/lBKoa9OAIOA}
\end{itemize}
 }

\bibliographystyle{IEEEtran}
\bibliography{egbib}

\end{document}